\PassOptionsToPackage{numbers,sort&compress}{natbib}
\documentclass{article}

\usepackage[preprint]{neurips_2026}

\usepackage[utf8]{inputenc} 
\usepackage[T1]{fontenc}    
\usepackage{hyperref}       
\usepackage{url}            
\usepackage{booktabs}       
\usepackage{amsfonts}       
\usepackage{nicefrac}       
\usepackage{microtype}      
\usepackage{xcolor}         

\usepackage{enumitem}
\usepackage{mathrsfs}   
\usepackage{amsmath,amssymb}
\usepackage{graphicx, wrapfig, float, indentfirst}
\usepackage{subcaption}
\usepackage[linesnumbered,ruled,vlined]{algorithm2e} 
\usepackage{float}
\usepackage{bbm}
\usepackage{pifont} 

\usepackage[table]{xcolor}
\usepackage{tabularx}
\usepackage{multirow}
\usepackage{makecell}
\usepackage{array}
\usepackage{booktabs,tabularx,array,multirow}
\newcolumntype{Y}{>{\centering\arraybackslash}X}
\usepackage{subcaption}  

\newcommand{\tb}{} 

\usepackage{etoc} 

\title{DictLLM: Post-training Compression of Large Language Model with Dictionary Kernels}

\author{%
  Jinho Park \\
  Columbia University \\
  \texttt{jp4327@columbia.edu} \\
  \And
  Se Young Chun \\
  Seoul National University \\
  \texttt{sychun@snu.ac.kr} \\
  \\
  \href{https://jp4327.github.io/dictllm/}{\texttt{jp4327.github.io/dictllm/}} \\
  \And
  Mingoo Seok \\
  Columbia University \\
  \texttt{ms4415@columbia.edu} \\
}

\begin{document}

\maketitle


\begin{abstract}
Efficient deployment of large language models (LLMs) on mobile devices requires balancing performance with compute and memory constraints. 
Post-training compression is an effective approach to managing this trade-off, while requiring much less computation than in-training counterparts.  
We propose DictLLM, a dictionary-based compression scheme in which weight matrices within grouped layers are decomposed into a shared dictionary and layer-specific sparse coefficients, directly obtained from a pre-trained model without further training.
Specifically, we cast DictLLM as a four-step post-training optimization.
First, we obtain shared dictionaries via singular value decomposition (SVD) of input-scaled weight matrices, amortizing rank across grouped layers; second, we sparsify the layer-specific coefficients to the target compression ratio using a dictionary-aware Hessian; and third, we refine the dictionaries given the resulting sparse coefficients.
Finally, we aggressively quantize both components using factor-specific Hessian information: input covariance for the dictionaries and the dictionary-aware Hessian for the coefficients.
The resulting representation reduces memory footprint by storing a single shared dictionary per layer group, along with sparse layer-specific coefficients.
On language modeling and zero-shot reasoning benchmarks, DictLLM consistently outperforms existing post-training low-rank compression methods and pushes the Pareto frontier between model size and performance.
DictLLM compresses the original LLaMA-7B model by 5.1\texttimes{} with only a 2.9-point increase in WikiText-2 perplexity, achieving a 3.6\texttimes{} higher compression ratio than the state-of-the-art low-rank method.

\end{abstract}    
\section{Introduction} \label{sec:intro}

Deploying large language models (LLMs) on mobile hardware is fundamentally limited by memory, not computation \cite{horowitz20141, chen2016eyeriss, williams2009roofline}.
Their billions of parameters exceed device capacity and dominate energy cost through repeated weight memory transfers.
Mobile devices do not have enough memory to store these models, and loading parameters to and from memory consumes significantly more energy than computations required for attention and feedforward (FFN) operations \cite{horowitz20141}.

In light of this, LLM compression has emerged as a promising approach for deploying LLMs on mobile graphics processing units (GPUs) with limited memory. 
In particular, post-training compression has attracted significant research attention, as it requires far fewer computational resources than its in-training counterparts. 
Model compression offers two key benefits: (1) it enables deployment of state-of-the-art LLMs on a mobile device, and (2) it alleviates memory-bandwidth bottlenecks, as characterized by the roofline model~\cite{jouppi2017datacenter,williams2009roofline}.
While compression also reduces the number of multiply–accumulate (MAC) operations, the dominant performance gains typically arise from reduced memory traffic rather than decreased computation~\cite{jouppi2017datacenter, lee2021link}.

Post-training model compression methods have sought to reduce model footprint along three independent axes: low-rank approximation shrinks matrix dimensions, pruning removes individual connections, and quantization lowers per-element bit width.
Low-rank approximation uses singular value decomposition (SVD)~\cite{eckart1936approximation} to factorize weight matrices into smaller ones~\cite{lee2024blast, wangsvd, wang2025basis, qinsidobi, yuan2023asvd, lu2026skipcat}.
However, meaningful compression requires the retained rank to be substantially smaller than the original matrix dimensions.
Since non-negligible spectral components can remain outside the retained subspace, aggressive low-rank approximation often degrades performance.
On the other hand, pruning \cite{han2015deep} allows sparsification of weight matrices while preserving their shape \cite{frantar2023sparsegpt, sun2024simple}.
Since neural processing units (NPUs) and GPUs support sparse matrix-matrix multiplication, pruned models can benefit from improved energy efficiency.
Orthogonal to these methods, quantization \cite{nvidia2024nvidia, egiazarian2026bridging, cheequip, frantar2023gptq, xiao2023smoothquant, ashkboos2024quarot, park2022bin, moon2026persasr} reduces the bit width of each element while keeping the number of model parameters constant.

\begin{figure*}[t] 
	\centering
	\includegraphics[width=0.95\linewidth]{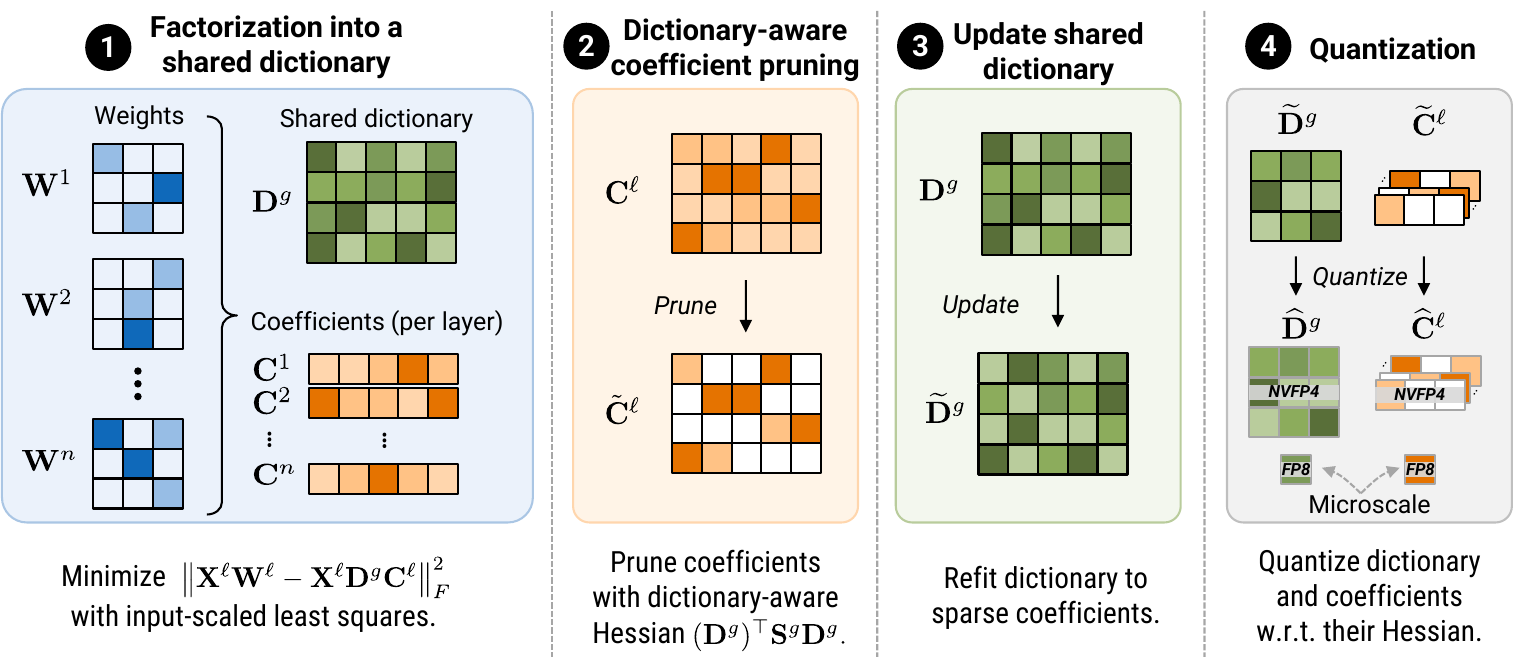}
    \vspace{-0.5em}
    \caption{\textbf{DictLLM: Dictionary-based compression method.}
    DictLLM resolves the weight memory bottleneck by compressing transformer weights into a shareable dictionary.
    It compounds four compression steps.
    \ding{202} The first step factorizes input-scaled weight matrices of grouped layers into a shared dictionary and layer-specific coefficients.
    \ding{203} We then prune the layer-specific coefficients using a dictionary-aware Hessian.
    \ding{204} We refit the shared dictionary to the resulting sparse coefficients.
    \ding{205} We quantize both the dictionary and the coefficients into NVFP4 using input statistics for the dictionary and dictionary-aware Hessians for the coefficients.
    The resulting compressed representation enables efficient projections by transferring a small set of shared dictionaries.
    }
	\label{fig:method}
    \vspace{-2em}
\end{figure*}

Parameter sharing~\cite{lanalbert,lou2022dictformer,bae2025relaxed,haydynamic} takes a different approach by reusing attention and feedforward weights across layers, enabling deeper models without additional parameters.
On the other hand, dictionary-based transformers relax strict sharing by representing weights as combinations of shared atoms and layer-specific coefficients~\cite{lou2022dictformer,moon2025t}.
This idea is closely related to classical dictionary learning, where algorithms such as the Method of Optimal Directions (MOD)~\cite{engan1999method} and \textit{k}-SVD~\cite{aharon2006k} reconstruct signals such as images from an overcomplete dictionary and sparse representation coefficients.
Although parameter sharing approaches can greatly reduce the number of parameters and memory traffic, they often require training the shared representation from scratch.
This in-training requirement of the method makes it difficult to apply directly as a post-training compressor for billion-scale LLMs.

Existing post-training methods \cite{lee2024blast, wangsvd, wang2025basis, qinsidobi, yuan2023asvd, lu2026skipcat, frantar2023sparsegpt, sun2024simple, nvidia2024nvidia, egiazarian2026bridging, cheequip, frantar2023gptq, xiao2023smoothquant, ashkboos2024quarot} and sharing-based architectures~\cite{lanalbert,lou2022dictformer,bae2025relaxed} therefore exploit complementary axes of redundancy, but they are typically applied in isolation.
A more effective post-training compressor should combine these benefits by sharing a reusable structure across layers while preserving layer-specific adaptation.
However, stacking these methods is nontrivial since each compression stage changes the weight and activation statistics used by the next stage, and a na\"ive composition can severely degrade performance.

We propose DictLLM, a post-training compression method that jointly achieves layer sharing, Hessian-aware sparsification, and 4-bit block floating-point (FP4) quantization via a single shared dictionary and sparse layer-specific coefficients within grouped layers (Fig.~\ref{fig:method}).
We develop a 3+1 approach that shows shared dictionary-based compression can be constructed via a three-step post-training process: input-scaled factorization, dictionary-aware Hessian coefficient pruning, and dictionary refitting.
This distinguishes our approach from prior work that constructs dictionary-based compression from scratch~\cite{lou2022dictformer} or requires principal component-based analysis and fine-tuning~\cite{zhussip2026share}.
In the last step, we show that the compressed model remains amenable to FP4~\cite{nvidia2024nvidia} quantization.

We highlight our contributions as follows:
\vspace{-0.5em} 
\begin{itemize}[leftmargin=3em]
    \setlength\parsep{0pt}
    
    \item We propose dictionary-based parameter sharing, which decomposes large weight matrices into shared dictionary kernels and layer-wise sparse representation coefficients.
    Our zero-shot \textit{post-training} compression method enables compression of a billion-parameter LLM.

    \item We combine activation-aware dictionary factorization and dictionary-induced Hessian pruning of the coefficients, then refit the shared dictionary to the sparse representation coefficients, improving reconstruction without changing sparsity or model size.
    
    \item We perform Hessian-aware FP4 quantization using the input covariance for the dictionary and the dictionary-based saliency for the coefficients.

    \item DictLLM reduces the original LLaMA-7B size by 5.1\texttimes{} with only a 2.9-point increase in WikiText-2 perplexity, improving over the state-of-the-art low-rank method by 3.6\texttimes{}.
\end{itemize}

\section{Related works} \label{sec:related}

\subsection{Post-training model compression}

\paragraph{Low-rank compression.}
Low-rank approximation is a natural approach for compressing transformer weights because attention and feedforward layers are dominated by large matrix multiplications.
Existing SVD methods reduce each weight matrix into two smaller factors, but aggressive rank truncation often causes substantial degradation in LLMs.
SVD-LLM~\cite{wangsvd} improves low-rank compression by using truncation-aware decomposition and activation information.
ASVD~\cite{yuan2023asvd} and Dobi-SVD~\cite{qinsidobi} also use activation-aware scaling to reduce the error from low-rank approximation.

\paragraph{Pruning.}
Pruning reduces model size by removing redundant weights while preserving the original matrix shapes.
Early pruning methods used magnitude or second-order saliency criteria to identify parameters to remove~\cite {lecun1989optimal, han2015deep}.
For LLMs, SparseGPT~\cite{frantar2023sparsegpt} introduced an efficient post-training pruning method that uses calibration activations as a Hessian proxy and compensates for pruning errors in a blockwise manner.
Wanda~\cite{sun2024simple} showed that combining weight magnitude with activation statistics can provide an effective pruning criterion.
These methods directly prune dense weight matrices~\cite{lee2026the}.
In contrast, our method prunes the layer-specific coefficient matrices after dictionary factorization and projects the activation covariance into the coefficient space induced by the shared dictionary.

\paragraph{Low-bit quantization.}
Quantization reduces memory footprint and bandwidth by representing model weights with fewer bits~\cite{lin2024awq, frantar2023gptq, ashkboos2024quarot, cheequip}.
Post-training quantization methods, such as GPTQ~\cite{frantar2023gptq}, use calibration activations to compensate for quantization errors without retraining.
Recent hardware-oriented formats further push weight precision below INT8 by using floating-point quantization with shared scales.
NVFP4~\cite{nvidia2024nvidia} represents groups of FP4 values with micro-scaling factors, enabling efficient low-bit inference on modern accelerators.
Microscaling FP4 methods have shown that careful scaling and error compensation are important for closing the gap between theoretical compression and practical accuracy~\cite{egiazarian2026bridging}.
DictLLM quantizes both the dictionary and the coefficients with NVFP4.
It compensates for dictionary quantization errors using the input covariance and for coefficient quantization errors using the dictionary-induced Hessian.

\subsection{Parameter sharing and dictionary learning}

\paragraph{Parameter sharing.}
Parameter sharing reduces model size by reusing weights across layers or recurrent computation steps.
ALBERT~\cite{lanalbert} shares transformer parameters across layers, showing that substantial parameter reduction is possible without changing the high-level architecture.
Deep equilibrium models~\cite{bai2019deep} take a related perspective by replacing explicit layer stacks with a recurrent computation that converges to a fixed point.
More recent methods relax strict sharing by adding layer-specific residual or low-rank components~\cite{bae2025relaxed,liaocompress,zhou2025large,zhaopivoting,li2026lestd}.

Basis Sharing~\cite{wang2025basis} further improves the compression trade-off by sharing basis matrices across layers instead of factorizing each layer independently.
DictLLM builds on this direction by recovering shared dictionaries via activation-aware factorization, while further sparsifying the layer-specific coefficients using a dictionary-aware Hessian and quantizing both factors with NVFP4.

\paragraph{Dictionary learning.}
Dictionary-based models represent weights or signals as combinations of shared atoms and sparse coefficients.
This idea is rooted in classical dictionary learning, where methods such as MOD~\cite{engan1999method} and \textit{k}-SVD~\cite{aharon2006k, makhov2025cospadi} alternate between sparse coding and dictionary updates to reconstruct signals from an overcomplete basis.
DictFormer~\cite{lou2022dictformer} applies this idea to transformers and substantially reduces model size by sharing dictionary atoms across layers.
T-REX~\cite{moon2025t} further shows that keeping the shared dictionary on chip can reduce external memory access and improve energy efficiency.
However, existing dictionary-based transformer approaches typically require training the dictionary representation from scratch, making them difficult to apply to billion-scale pretrained LLMs.
DictLLM instead constructs a dictionary representation, prunes coefficients with a dictionary-induced Hessian, and refits the dictionary to the sparse coefficients.
    
\section{Proposed methods} \label{sec:method}

\begin{figure}[t]
    \centering
    \begin{minipage}[t]{0.48\textwidth}
        \centering
        \includegraphics[width=\linewidth]{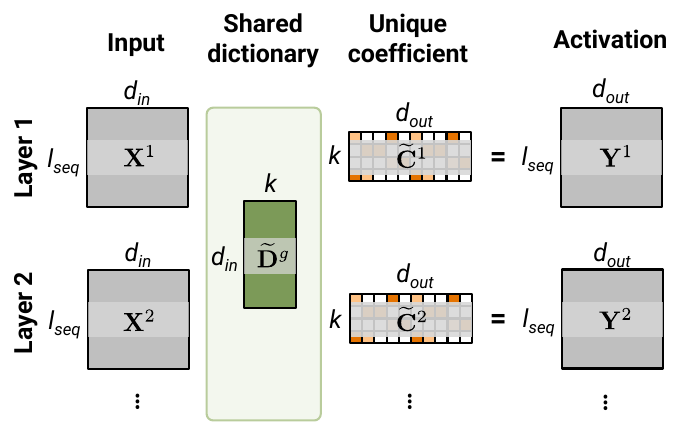}
        \caption{\textbf{Dictionary-based forward operation.} 
        Each weight matrix $\mathbf{W}^{\ell}$ is replaced by the product of a shared dictionary $\widetilde{\mathbf{D}}^{g}$ and a layer-specific sparse coefficient matrix $\widetilde{\mathbf{C}}^{\ell}$. The forward pass computes $\mathbf{Y}^\ell = \mathbf{X}^{\ell}\widetilde{\mathbf{D}}^{g}\widetilde{\mathbf{C}}^{\ell}$. 
        }
        \label{fig:dict_forward}
    \end{minipage}
    \hfill
    \begin{minipage}[t]{0.48\textwidth}
        \centering
        \includegraphics[width=\linewidth]{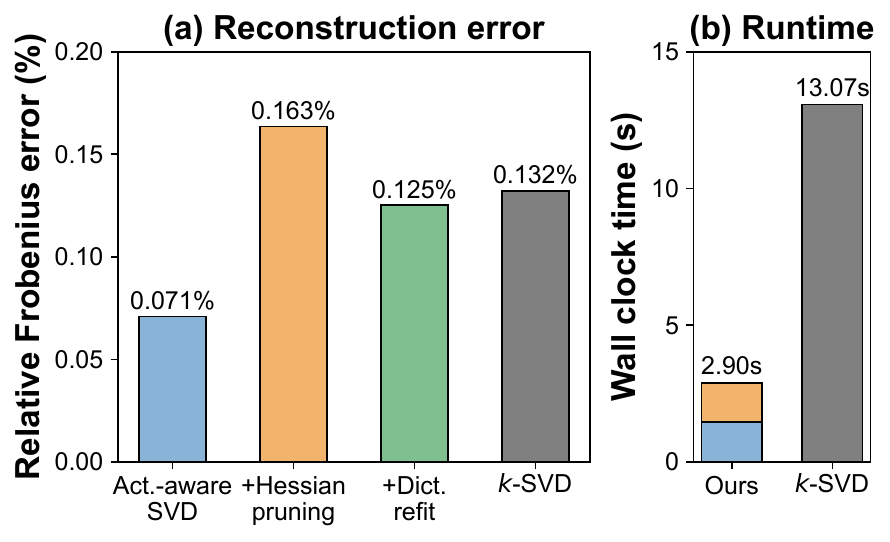}
        \caption{\tb
        \textbf{Motivation for staged dictionary refinement.}
        We compare (a) activation reconstruction error and (b) runtime of our factorize--prune--update pipeline against \textit{k}-SVD on a single LLaMA-7B query projection.
        }
        \label{fig:motivation}
    \end{minipage}
    \vspace{-1em}
\end{figure}

\subsection{Shared dictionary representation}
\label{subsec:dict_rep}

We first describe the shared-dictionary representation and the forward operation.
For a layer group $g$ with layers $\ell \in g$ that share a dictionary, we define input dimension $d_\text{in}$, output dimension $d_\text{out}$, and dictionary size $k$ (Fig.~\ref{fig:dict_forward}).
A shared dictionary $\widetilde{\mathbf{D}}^{g} \in \mathbb{R}^{d_\text{in} \times k}$ is reused by every layer in the group, while the coefficient matrix $\widetilde{\mathbf{C}}^{\ell} \in \mathbb{R}^{k \times d_\text{out}}$ of each layer is not shared.
The original weight matrix $\mathbf{W}^{\ell} \in \mathbb{R}^{d_\text{in} \times d_\text{out}}$ of layer $\ell$ is approximated column-wise as
\begin{equation}
    \widetilde{\mathbf{W}}^{\ell}_{:,j} = \sum_{m \in \mathcal{S}^{\ell}_j} \widetilde{\mathbf{D}}_{:,m}^{g}\, \widetilde{\mathbf{C}}^{\ell}_{m,j},
    \label{eq:forward}
\end{equation}
where $\mathcal{S}^{\ell}_j \subseteq [k]$ denotes the indices of active dictionary atoms. 
The coefficient sparsity is defined as $p = 1 - |\mathcal{S}^{\ell}_j|/k$.
Equation~\ref{eq:forward} can be written in matrix form as $\widetilde{\mathbf{W}}^{\ell} = \widetilde{\mathbf{D}}^{g} \widetilde{\mathbf{C}}^{\ell}$, where $\widetilde{\mathbf{C}}^{\ell}$ is sparse.


\begin{figure}[t]
    \centering
    \begin{minipage}[t]{0.5\textwidth}
    \IncMargin{1.1em}
    \begin{algorithm}[H]
        \SetNlSty{normaltext}{}{:}
        \caption{Dictionary-based compression}
        \label{alg:dict}
    
        \KwIn{Pretrained weights $\{\mathbf{W}_{t}^{\ell}\}_{\ell=1}^{L}$, 
            grouped layer index sets $\mathcal{G}$, 
            calibration set $\mathcal{X}$, sparsity $p$}
        \KwOut{Compressed parameters $\{\widehat{\mathbf{D}}, \widehat{\mathbf{C}}\}$}
    
        $\{\mathbf{S}_{t}^{\ell}\}_{\ell,t}
        \leftarrow
        \operatorname{CollectCovariance}(f_{\mathbf{W}}, \mathcal{X})$\nllabel{line:collect-cov}\;
        
        \ForEach{weight type $t$}
        {
            \ForEach{layer group $g \in \mathcal{G}^{t}$}
            {
                $\mathcal{W}_{t}^{g} \leftarrow \{\mathbf{W}_{t}^{\ell}: \ell \in g\}$\;
                $\mathbf{S}_{t}^{g} \leftarrow \sum_{\ell \in g}\mathbf{S}_{t}^{\ell}$\;
                $k_{t}^{g} \leftarrow \operatorname{Rank}(\mathcal{W}_{t}^{g})$\;
        
                $\mathbf{D}_{t}^{g}, \{\mathbf{C}_{t}^{\ell}\}_{\ell \in g}
                \leftarrow
                \operatorname{Factorize}(\mathcal{W}_{t}^{g}, \mathbf{S}_{t}^{g}, k_{t}^{g})$\;
        
                $\mathbf{H}_{t}^{\mathbf{C},g}
                \leftarrow
                (\mathbf{D}_{t}^{g})^{\top}\mathbf{S}_{t}^{g}\mathbf{D}_{t}^{g}$\;
        
                \ForEach{$\ell \in g$}
                {
                    $\widetilde{\mathbf{C}}_{t}^{\ell}
                    \leftarrow
                    \operatorname{Prune}(\mathbf{C}_{t}^{\ell}, \mathbf{H}_{t}^{\mathbf{C},g}, p)$\;
                }
                $\widetilde{\mathbf{D}}_{t}^{g}
                \leftarrow
                \operatorname{Refit}(\mathcal{W}_{t}^{g}, \{\widetilde{\mathbf{C}}_{t}^{\ell}\}_{\ell \in g})$\;
        
                $\widehat{\mathbf{D}}_{t}^{g}
                \leftarrow
                \operatorname{Quantize}(\widetilde{\mathbf{D}}_{t}^{g}, \mathbf{S}_{t}^{g})$\;
                
                \ForEach{$\ell \in g$}
                {
                    $\widehat{\mathbf{C}}_{t}^{\ell}
                    \leftarrow
                    \operatorname{Quantize}(\widetilde{\mathbf{C}}_{t}^{\ell}, \mathbf{H}_{t}^{\mathbf{C},g})$\;
                }
            }
        }
    \end{algorithm}
    \end{minipage}
    \hfill
    \begin{minipage}[t]{0.49\textwidth}
    \IncMargin{1.1em}

    \begin{algorithm}[H]
        \SetNlSty{normaltext}{}{:}
        \caption{Dictionary-aware coefficient pruning}
        \label{alg:dict_prune}
        
        \KwIn{Coefficient matrix $\mathbf{C}$,
              dictionary $\mathbf{D}$, input covariance $\mathbf{S}$,
              sparsity $p$, damping $\lambda$, block size $B$}
        \KwOut{Pruned coefficient matrix $\widetilde{\mathbf{C}}$}
        
        $\mathbf{H}^{-1}
        \leftarrow
        (\mathbf{D}^{\top}\mathbf{S}\mathbf{D} + \lambda\mathbf{I})^{-1}$\;
        $\mathbf{d} \leftarrow \operatorname{diag}(\mathbf{H}^{-1})$\;
        $\widetilde{\mathbf{C}} \leftarrow \mathbf{C}$\;
        
        \For{$j \leftarrow 0,\,B,\,2B,\,\ldots$}
        {
            $\mathbf{R}_{:,\,j:j+B}
            \leftarrow
            \operatorname{SaliencyMap}(\widetilde{\mathbf{C}}_{:, j:j+B}, \mathbf{d})$\;
        
            \For{$i \leftarrow 0$ \KwTo $\lfloor p\cdot k \rfloor - 1$}
            {
                $\mathbf{c}
                \leftarrow
                \mathbf{R}_{i,\,j:j+B}$\;
                \tcp{Least salient rows}
        
                $\mathbf{e}
                \leftarrow
                \widetilde{\mathbf{C}}_{[\mathbf{c},\,j:j+B]}
                \odot
                \mathbf{d}_{[\mathbf{c}]}^{-1}$\;
                \tcp{Corresponding error}
        
                $\widetilde{\mathbf{C}}_{:,\,j:j+B}
                \leftarrow
                \widetilde{\mathbf{C}}_{:,\,j:j+B}
                -
                \mathbf{H}^{-1}_{:,[\mathbf{c}]}\operatorname{diag}(\mathbf{e})$\;
            }
        
            $\widetilde{\mathbf{C}}_{:,\,j:j+B}
            \leftarrow
            \widetilde{\mathbf{C}}_{:,\,j:j+B}
            \odot
            \mathbf{M}_{:,\,j:j+B}$
            \tcp{Apply pruning}
        }
    
    \end{algorithm}
    \end{minipage}
\end{figure}

\subsection{Dictionary-based compression objective}

We compute the dictionary and sparse coefficients sequentially because joint optimization would require solving a large discrete-continuous problem over shared factors and sparsity masks.
DictLLM solves an activation-weighted reconstruction problem,
\begin{equation}
    \min_{\mathbf{D}^{g},\,\{\mathbf{C}^{\ell}\}_{\ell \in g}}
    \sum_{\ell \in g}
    \left\|
     \mathbf{X}^{\ell} \mathbf{W}^{\ell}
    -
    \mathbf{X}^{\ell} \mathbf{D}^{g} \mathbf{C}^{\ell}
    \right\|_{F}^{2}
    \label{eq:activation_weighted_factorization}
\end{equation}
where $\|\cdot\|_F$ denotes the Frobenius norm.
Algorithm~\ref{alg:dict} delineates the decomposition steps.
We first decompose grouped model weights into a shared dictionary $\mathbf{D}^{g}$ and dense layer-specific coefficients $\{\mathbf{C}^{\ell}\}_{\ell \in g}$ using activation-aware low-rank factorization.
Next, we sparsify the coefficients using a dictionary-aware Hessian criterion with the dictionary fixed.
The final step refits the shared dictionary based on the resulting coefficients.

\paragraph{{\tb Motivation for staged dictionary refinement.}}
To motivate our post-training approximation, we compare our method, consisting of activation-aware SVD, Hessian coefficient pruning, and post-pruning dictionary update, against $k$-SVD with an undercomplete dictionary ($k<d_\text{in}$) on the first layer of LLaMA-7B query projection (Fig.~\ref{fig:motivation}).
\textit{k}-SVD is an alternative to our sequential pipeline for learning a sparse dictionary representation from scratch.
However, full sparse dictionary learning is expensive at the LLM scale because it requires repeated atom-wise and coefficient updates over a large number of weight columns.
We measure the relative activation reconstruction error $\|\mathbf{X}\mathbf{W}-\mathbf{X}\widetilde{\mathbf{W}}\|_{F}/
\|\mathbf{X}\mathbf{W}\|_{F}$ using 256 calibration samples from WikiText-2.
Our three-stage pipeline achieves lower reconstruction error than \textit{k}-SVD while running 3.4\texttimes{} faster.
We discuss the formulation of \textit{k}-SVD in Appendix~\ref{app:ksvd_relation}.

\subsection{Dictionary-based compression pipeline: factorize, prune, refit, and quantize}

\paragraph{Step 1: Activation-aware dictionary factorization.}
We first factorize each group of weights into a shared dictionary and layer-specific coefficients using activation-aware SVD.
For a layer group $g=\{\ell_1,\ldots,\ell_n\}$, we first obtain the concatenated weight $\mathbf{W}^{g}=[\mathbf{W}^{\ell_1}|\mathbf{W}^{\ell_2}|\cdots|\mathbf{W}^{\ell_n}]$ for $n$ weights in $g$, similar to \cite{wang2025basis}.
Given calibration examples $\mathcal{X}$, we run the original model \(f_{\mathbf W}\) to collect each linear module’s input activations \(\mathbf X^\ell\) and aggregate it as $\mathbf{S}^{g}=\sum_{\ell \in g}\mathbf{S}^{\ell}=\sum_{\ell \in g}(\mathbf{X}^{\ell})^{\top}\mathbf{X}^{\ell} \in \mathbb{R}^{d_\text{in} \times d_\text{in}}$.
This is equivalent to computing the covariance after vertically concatenating the inputs from the grouped layers.
We then perform rank-$k$ SVD on the input-scaled concatenated weight $(\mathbf{S}^{g})^{1/2}\mathbf{W}^{g}\approx \mathbf{U}_k \mathbf{\Sigma}_k \mathbf{V}^{\top}_k$, where $\mathbf{U}_k \in \mathbb{R}^{d_\text{in} \times k}$, $\mathbf{\Sigma}_k \in \mathbb{R}^{k \times k}$, and $\mathbf{V}_k \in \mathbb{R}^{nd_\text{out} \times k}$.
The shared dictionary is attained as $\mathbf{D}^{g}=(\mathbf{S}^{g})^{-1/2}\mathbf{U}_k\mathbf{\Sigma}_k$, and $\mathbf{V}^{\top}_k$ is equivalent to layer-specific coefficients $\{\mathbf{C}^{\ell}\}_{\ell\in g}$.
Appendix \ref{app:compression_param} includes details on rank $k$ in relation to the compression ratio.

\paragraph{Step 2: Dictionary-aware coefficient pruning.}
After dictionary factorization, we prune the layer-specific coefficients $\{\mathbf{C}^{\ell}\}_{\ell\in g}$, keeping the shared dictionary $\mathbf{D}^{g}$ fixed.
Unlike SparseGPT~\cite{frantar2023sparsegpt}, which prunes the dense weight $\mathbf{W}^{\ell}$ using the input covariance $\mathbf{S}^{\ell}=({\mathbf{X}^{\ell}})^{\top}\mathbf{X}^{\ell} \in \mathbb{R}^{d_\text{in} \times d_\text{in}}$, DictLLM prunes in the coefficient space defined by the shared dictionary.
Pruning the coefficient induces a structured weight perturbation $\Delta\mathbf{W}^{\ell}=\mathbf{D}^{g}\Delta\mathbf{C}^{\ell}$.
The local reconstruction error from coefficient pruning is
\begin{equation}
    \|\mathbf{X}^{\ell}\mathbf{D}^{g}\Delta\mathbf{C}^{\ell}\|_{F}^{2}.
\end{equation}
This yields the dictionary-aware coefficient Hessian $\mathbf{H}^{\mathbf{C},g}=(\mathbf{D}^{g})^{\top}\mathbf{S}^{g}\mathbf{D}^{g}\in\mathbb{R}^{k\times k}$, where $\mathbf{S}^{g}$ is the grouped input covariance from calibration examples.
We derive this coefficient-space Hessian from the local reconstruction objective in Appendix~\ref{app:coef_hessian}.

Algorithm~\ref{alg:dict_prune} summarizes the pruning procedure.
$\operatorname{SaliencyMap}(\widetilde{\mathbf{C}}_{:,j:j+B}, \mathbf{d})$ returns $\mathbf{R}_{:,j:j+B} \in \mathbb{Z}^{k \times B}$ for each block of $B$ columns.
Each column of $\mathbf{R}$ sorts the $k$ row indices by ascending saliency calculated by $\widetilde{\mathbf{C}}_{i,j}^{2}/\mathbf{d}_{i}$ where $\mathbf{d} \in \mathbb{R}^{k}$.
At pruning step $i$, we find the least salient row index per column for the $B$ columns in the block as $\mathbf{c} \in \mathbb{Z}^{B}$, and compute the error used to compensate the remaining rows as $\mathbf{e} \in \mathbb{R}^{B}$.
After $\lfloor p \cdot k \rfloor$ steps, the binary mask $\mathbf{M}$ prunes the rows indexed by $\mathbf{R}_{0:\lfloor p \cdot k \rfloor,:}$.
The key efficiency gain comes from operating in the coefficient space.
Since $k < d_\text{in}$, inverting $\mathbf{H}^{\mathbf{C},g} \in \mathbb{R}^{k \times k}$ costs $\mathcal{O}(k^{3})$ rather than $\mathcal{O}(d_\text{in}^{3})$ of ~\cite{lecun1989optimal,frantar2023sparsegpt}; the inverse is shared across all coefficient columns, reducing each column update to $\mathcal{O}(k^{2})$.

\paragraph{Step 3: Updating shared dictionary with sparse coefficients.}
We refit the dictionary to the sparse coefficients while keeping the coefficient values and sparsity pattern fixed.
This is necessary because the shared dictionary $\mathbf{D}^{g}$ is obtained before pruning and is therefore optimized for dense coefficients, not for the sparse support induced by $\widetilde{\mathbf{C}}^{\ell}$.
We update the shared dictionary
$
    \mathbf{D}^{g}
    \leftarrow
    \mathbf{W}^{g}(\widetilde{\mathbf{C}}^{g})^{\top}
    (
    \widetilde{\mathbf{C}}^{g}(\widetilde{\mathbf{C}}^{g})^{\top}
    + \lambda \mathbf{I}
    )^{-1}
$
where
$\widetilde{\mathbf{C}}^{g}=[\widetilde{\mathbf{C}}^{\ell_1}|\cdots|\widetilde{\mathbf{C}}^{\ell_n}]$ for a layer group $g$.
This is a solution to a least-squares problem,
$
    \min_{\mathbf{D}^{g}}
    ||\mathbf{X}^g(
    \mathbf{W}^{g}
    -
    \mathbf{D}^{g}\widetilde{\mathbf{C}}^{g}
    )||_{F}^{2}
$
where
$\mathbf{X}^{g}
=
[(\mathbf{X}^{\ell_1})^{\top},
\ldots,
(\mathbf{X}^{\ell_n})^{\top}]^{\top}$
denotes the vertical concatenation of calibration activations.
Hence, 
$(\mathbf{X}^{g})^{\top}\mathbf{X}^{g}
=
\sum_{\ell\in g}(\mathbf{X}^{\ell})^{\top}\mathbf{X}^{\ell}
=
\mathbf{S}^{g}$.
This update improves reconstruction under fixed sparse support without changing coefficient sparsity.

\paragraph{Step 4: Hessian-aware NVFP4 quantization.}
We quantize the dictionary and coefficients using respective Hessians.
For the dictionary $\mathbf{D}^{g}$, we group input covariance as the Hessian since the module input is the original activation.
For the coefficient $\mathbf{C}^{\ell}$, the Hessian proxy is
$\mathbf{H}^{\mathbf{C},g}=(\mathbf{D}^{g})^{\top}\mathbf{S}^{g}\mathbf{D}^{g}$,
since the module input is $\mathbf{X}^{\ell}\mathbf{D}^{g}$.
Thus, coefficient quantization uses the same dictionary-induced Hessian as coefficient pruning.

With these Hessians, we apply error compensation and quantize the dictionary and coefficients using NVFP4.
During coefficient quantization, we preserve the pruning mask by fixing masked entries to zero and compensating for the resulting errors.
We denote the NVFP4 quantizer as $\mathcal{Q}_{\mathrm{NVFP4}}(\cdot)$ and obtain the final compressed dictionary as $\widehat{\mathbf{D}}^{g}=\mathcal{Q}_{\mathrm{NVFP4}}(\mathbf{D}^{g})$ and coefficient $\widehat{\mathbf{C}}^{\ell}=\mathcal{Q}_{\mathrm{NVFP4}}(\widetilde{\mathbf{C}}^{\ell})$.
Following the NVFP4 format~\cite{nvidia2024nvidia}, each block of values is represented by low-precision FP4 together with a shared microscale.    
\section{Experiments} \label{sec:experiments}

\subsection{Experimental setup}

\paragraph{Models and datasets.}
We use DictLLM on LLaMA-7B and LLaMA2-7B~\cite{touvron2023llama}, OPT-6.7B~\cite{zhang2022opt}, and Mistral-7B~\cite{jiang2023mistral} using the HuggingFace checkpoint.
We evaluate language modeling performance on WikiText-2~\cite{merity2017pointer}, Penn TreeBank (PTB)~\cite{marcus1993building}, and Colossal Clean Crawled Corpus (C4)~\cite{raffel2020exploring}.
We also evaluate zero-shot reasoning performance on seven benchmarks: OpenBookQA~\cite{mihaylov2018can}, WinoGrande~\cite{sakaguchi2020winogrande}, HellaSwag~\cite{zellers2019hellaswag}, PIQA~\cite{bisk2020piqa}, MathQA~\cite{amini2019mathqa}, ARC-Easy, and ARC-Challenge~\cite{clark2018think}.
All reasoning evaluations are performed in the zero-shot setting using the LM-Evaluation-Harness~\cite{eval-harness}.

\paragraph{Baselines and metrics.}
We compare DictLLM against post-training low-rank compression methods.
ASVD~\cite{yuan2023asvd} applies activation-aware scaling before low-rank decomposition.
SVD-LLM~\cite{wangsvd} improves low-rank compression with truncation-aware singular value decomposition.
Basis Sharing~\cite{wang2025basis} shares low-rank bases across transformer layers and is the closest baseline to our dictionary factorization.
We report perplexity on language modeling tasks and accuracy for reasoning tasks.
We measure compression gain based on the number of non-zero entries in the dictionary and coefficient matrices.

\paragraph{Implementation details.}
We use models with checkpoints from HuggingFace Transformers.
We collect calibration statistics from 256 WikiText-2 samples with a sequence length of 2048.
All experiments are run on dual NVIDIA RTX PRO 6000 Blackwell GPUs.
We apply NVFP4 quantization post hoc to both dictionary and coefficient factors based on QuTLASS~\cite{qutlass2025}.

\subsection{Experimental results}

\newcommand{\sysname}{{DictLLM}\xspace}
\newcommand{\tgr}{\color{gray}}

\begin{table*}[t]
\centering
\captionof{table}{
\textbf{Benchmark on language modeling and reasoning tasks.}
We benchmark DictLLM against low-rank compression baselines on language modeling and commonsense reasoning tasks for LLaMA-7B at different compression ratios.
We additionally report the FP4-quantized DictLLM\textsubscript{FP4} models at the bottom of the table.
We measure perplexity on WikiText-2, Penn TreeBank (PTB), and C4, and zero-shot accuracy on OpenBookQA (OB), WinoGrande (WG), HellaSwag (HS), PIQA (PQ), MathQA (MQ), ARC-Easy (AE), and ARC-Challenge (AC).
}
\resizebox{1\textwidth}{!}{
\begin{tabular}{c|c|ccc|cccccccc}
    \toprule
    {\textbf{Comp.}} & {\multirow{2}{*}{\textbf{Method}}}  & \multicolumn{3}{c|}{\textbf{Perplexity} $\downarrow$}       & \multicolumn{8}{c}{\textbf{Zero-shot accuracy} $\uparrow$}        \\
    {\textbf{gain}}                      &  {}                       & WikiText-2 & PTB & C4   & OB & WG & HS& PQ & MQ  & AE & AC & {Average}        \\
    \midrule
    \midrule

    {\tgr 1\texttimes{}}  & {\tgr Original}  & {\tgr 5.68}   & {\tgr 8.35}     & {\tgr 7.34}      & {\tgr 0.28} & {\tgr 0.67} & {\tgr 0.56}  & {\tgr 0.78} & {\tgr 0.27}  & {\tgr 0.67}  & {\tgr 0.38} & {\tgr 0.52}     \\ \midrule
    
    {\multirow{4}{*}{1.25\texttimes{} }} & {ASVD \cite{yuan2023asvd}} & 11.14          & 16.55         & 15.93           & 0.25  & 0.64  & 0.41 & 0.68 & 0.24 & 0.53 & 0.27 & 0.43 \\
    {}                      & {SVD-LLM \cite{wangsvd}}              & 7.94           & 18.05         & 15.93           & 0.22  & 0.63  & 0.43 & 0.69 & 0.24 & 0.58 & 0.29 & 0.44 \\
    {}                      & {Basis Sharing \cite{wang2025basis}}  & 7.74           & 17.35         & 15.03           & 0.28  & 0.66  & 0.46 & 0.71 & 0.25 & 0.66 & \textbf{0.36} & 0.48 \\
    {} & {\sysname (Ours)} & \textbf{6.89} & \textbf{12.27} & \textbf{11.75} & \textbf{0.33} & \textbf{0.67} & \textbf{0.50} & \textbf{0.74} & \textbf{0.26} & \textbf{0.69} & \textbf{0.36} & \textbf{0.51} \\
    
    \midrule

    {\multirow{4}{*}{1.43\texttimes{} }} & {ASVD \cite{yuan2023asvd}} & 51             & 70            & 41              & 0.18  & 0.53  & 0.37 & 0.65 & 0.21 & 0.43 & 0.25 & 0.38 \\
    {}                      & {SVD-LLM \cite{wangsvd}}              & 9.56           & 29.44         & 25.11           & 0.20  & 0.59  & 0.40 & 0.65 & 0.22 & 0.48 & 0.26 & 0.40 \\
    {}                      & {Basis Sharing \cite{wang2025basis}}  & 9.25           & 29.12         & 22.46           & 0.27  & 0.63  & 0.40 & 0.68 & 0.24 & 0.63 & 0.30 & 0.45 \\
    {} & {\sysname (Ours)} & \textbf{7.79} & \textbf{17.62} & \textbf{15.20} & \textbf{0.29} & \textbf{0.66} & \textbf{0.46} & \textbf{0.71} & \textbf{0.24} & \textbf{0.66} & \textbf{0.35} & \textbf{0.48} \\

    \midrule
    
    {\multirow{4}{*}{1.67\texttimes{} }} & {ASVD \cite{yuan2023asvd}} & 1407           & 3292          & 1109            & 0.13  & 0.48  & 0.26 & 0.55 & 0.19 & 0.28 & 0.22 & 0.30 \\ 
    {}                      & {SVD-LLM \cite{wangsvd}}              & 13.11          & 63.75         & 49.83           & 0.19  & 0.58  & 0.33 & 0.60 & 0.21 & 0.42 & 0.25 & 0.37 \\
    {}                      & {Basis Sharing \cite{wang2025basis}}  & 12.39          & 55.78         & 41.28           & 0.22  & 0.61  & 0.35 & 0.62 & 0.23 & 0.52 & 0.27 & 0.40 \\ 
    {} & {\sysname (Ours)} & \textbf{9.37} & \textbf{29.75} & \textbf{23.08} & \textbf{0.27} & \textbf{0.64} & \textbf{0.40} & \textbf{0.68} & \textbf{0.23} & \textbf{0.63} & \textbf{0.30} & \textbf{0.45} \\

    \midrule
    
    {\multirow{4}{*}{2\texttimes{} }} & {ASVD \cite{yuan2023asvd}}  & 15358          & 47690         & 27925           & 0.12  & 0.51  & 0.26 & 0.52 & 0.19 & 0.26 & 0.22 & 0.30 \\ 
    {}                      & {SVD-LLM \cite{wangsvd}}              & 23.97          & 150.58        & 118.57          & 0.16  & 0.54  & 0.29 & 0.56 & 0.21 & 0.33 & 0.23 & 0.33 \\
    {}                      & {Basis Sharing \cite{wang2025basis}}  & 19.99          & 126.35        & 88.44           & 0.18  & 0.57  & 0.31 & 0.58 & 0.22 & 0.42 & 0.23 & 0.36 \\
    {} & {\sysname (Ours)} & \textbf{10.59} & \textbf{35.67} & \textbf{29.12} & \textbf{0.24} & \textbf{0.63} & \textbf{0.38} & \textbf{0.65} & \textbf{0.23} & \textbf{0.61} & \textbf{0.29} & \textbf{0.43} \\

    \midrule
    \midrule

    {5\texttimes{} } & {\multirow{3}{*}{\makecell{DictLLM\textsubscript{FP4} \\ (Ours)}}}  & 8.48 & 21.40 & 17.04 & 0.29 & 0.65 & 0.44 & 0.70 & 0.24 & 0.64 & 0.33 & 0.47 \\

    6\texttimes{} & {} & 10.37 & 35.16 & 27.07 & 0.26 & 0.61 & 0.38 & 0.66 & 0.23 & 0.59 & 0.28 & 0.43 \\

    7\texttimes{} & {} & 11.92 & 43.14 & 33.90 & 0.24 & 0.59 & 0.35 & 0.63 & 0.22 & 0.54 & 0.26 & 0.40 \\
    
    \bottomrule
\end{tabular}
}
\vspace{-1em}
\label{tab:benchmark}
\end{table*}

\paragraph{Benchmark on generation and reasoning datasets.}
We report language modeling perplexity and zero-shot reasoning accuracy in Table~\ref{tab:benchmark}.
DictLLM consistently improves over low-rank baselines~\cite{wang2025basis,wangsvd} across compression ratios from 1.25\texttimes{} to 2\texttimes{}.
For example, DictLLM achieves an average zero-shot accuracy of 0.43 at 2\texttimes{} compression, compared with 0.40 for Basis Sharing at 1.67\texttimes{} compression.
We also report FP4-quantized variants, DictLLM\textsubscript{FP4}, at the bottom of Table~\ref{tab:benchmark}.
Notably, DictLLM\textsubscript{FP4} achieves an average accuracy of 0.40 at 7\texttimes{} compression, outperforming Basis Sharing at 2\texttimes{} compression, which achieves 0.36 average accuracy.
We provide the sparsity level to reach the desired compression ratio in Table~\ref{tab:supp-sparsity-llama}.

\begin{wrapfigure}{o}{0.6\textwidth}
    \vspace{-2.1em}
    \centering
    \includegraphics[width=1.0\linewidth]{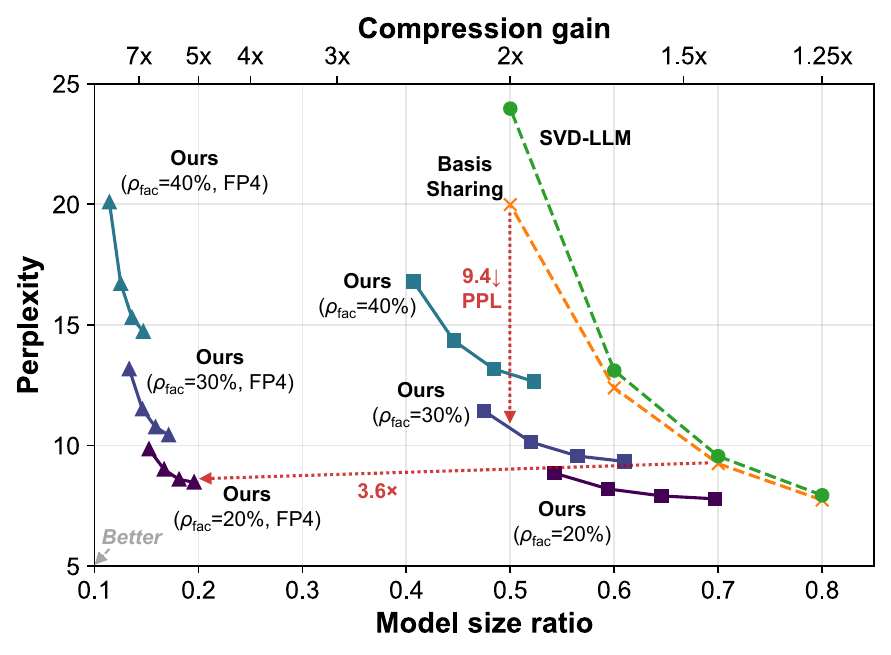}
    \vspace{-1em}
    \caption{\textbf{Perplexity-model size trade-off.}
        WikiText-2 perplexity of compressed LLaMA-7B across factorization ratios and coefficient sparsities.
        DictLLM factorizes weights at rate $\rho_{\mathrm{fac}}$ and then prunes layer-specific coefficients with 20\%--50\% sparsity.
        We also report their NVFP4 versions.
        }
    \vspace{-1.5em}
    \label{fig:PM_trade}
\end{wrapfigure}

\paragraph{Perplexity-model size trade-off.}
We evaluate the trade-off between WikiText-2 perplexity and model size ratio in Fig.~\ref{fig:PM_trade}.
DictLLM achieves a 5.1\texttimes{} compression ratio on the original LLaMA model with only a 2.9-point increase in WikiText-2 perplexity, providing 3.6\texttimes{} higher compression than the state-of-the-art low-rank method.
In addition, DictLLM reduces perplexity by 9.4 points compared to Basis Sharing~\cite{wang2025basis} with 2\texttimes{} compression on WikiText-2.
DictLLM provides a continuous compression curve by controlling the parameter count via two knobs: dictionary size and coefficient sparsity.
As the target model size decreases, perplexity gradually increases, reflecting stronger compression applied to both shared dictionaries and layer-specific coefficients.
We first determine the rank $k$ from the factorization compression rate $\rho_{\mathrm{fac}}$, then apply 20\%--50\% sparsity to the per-layer coefficients.
The final compression gain is therefore determined jointly by $\rho_{\mathrm{fac}}$ and the coefficient sparsity $p$.

\begin{table}[t]
    \centering
    \small
    \begin{minipage}[t]{0.49\textwidth}
        \centering
        \caption{
        \textbf{Ablation of pruning target.}
        We ablate the pruning target with sparsity $p$: the dictionary $\mathbf{D}$, its coefficients $\mathbf{C}$, or both.
        Perplexity is measured on WikiText-2 using LLaMA-7B, yielding an overall compression gain of 1.43\texttimes {}.
        }
        \label{tab:prune_target_ablation}
        \vspace{1em}
        \begin{tabular}{ccc}
            \toprule
            Dict. $\mathbf{D}$ ($p$) & Coef. $\mathbf{C}$ ($p$) & Perplexity $\downarrow$ \\
            \midrule
            Dense (0\%)     & Dense (0\%)       & 9.25          \\
            Dense (0\%)     & Sparse (19.4\%)   & \textbf{7.83} \\
            Sparse (35.0\%) & Dense (0\%)       & 9.04          \\
            Sparse (12.5\%) & Sparse (12.5\%)   & 7.99          \\
            \bottomrule
        \end{tabular}
    \end{minipage}
    \hfill
    \begin{minipage}[t]{0.49\textwidth}
        \centering
        \caption{
        \textbf{Choice of pruning method.} 
        We prune the coefficients of LLaMA-7B based on magnitude, full Hessian inversion, and Hessian with lazy updates.
        We measure the perplexity on WikiText-2 with 50\% sparsity on a 20\% factorized model.
        }
        \label{tab:pruning_method}
        \vspace{1em}
        \begin{tabular}{ccc}
            \toprule
            {Method}               & Perplexity $\downarrow$ & Time $\downarrow$ \\
            \midrule
            Magnitude           & 47.43             & \textbf{44.6 min} \\
            Full inversion      & \textbf{9.89}     & 13.49 h \\
            Lazy update (Ours)   & \textbf{9.89}    & 55.6 min \\
            \bottomrule
        \end{tabular}
    \end{minipage}
    \vspace{-1em}
\end{table}

\paragraph{Ablation of pruning target.}
Table~\ref{tab:prune_target_ablation} examines whether sparsity should be applied to the shared dictionary, the layer-specific coefficients, or both.
We control the sparsity of the pruning target to reach a 30\% compression ratio.
The most interesting finding is that pruning the coefficients gives the lowest perplexity of all combinations.
This suggests that the dictionary stores shared information reused across layers, whereas the coefficients exhibit more layer-specific redundancy.
Therefore, DictLLM keeps the dictionary dense and applies sparsity only to the coefficients.

\paragraph{Choice of pruning method.}
We compare three coefficient pruning strategies to isolate the effect of dictionary-aware Hessian compensation (Table~\ref{tab:pruning_method}).
Magnitude pruning is fast but ignores activation sensitivity and the dictionary, leading to significant degradation in perplexity.
Full Hessian inversion gives much better perplexity by compensating for pruning errors, but its cost is impractical for repeated compression.
Our dictionary-aware lazy-update method preserves the same curvature information while avoiding the expensive full inverse update.
As shown in Table~\ref{tab:pruning_method}, it matches the perplexity of the full inversion method while being 14.6\texttimes{} faster, with a duration similar to the magnitude-based method.

\begin{wrapfigure}{r}{0.4\textwidth}
    \vspace{-2em}
    \centering
    \includegraphics[width=1.0\linewidth]{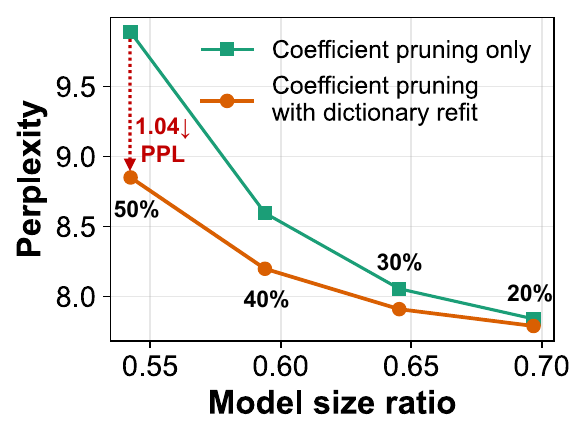}
    \vspace{-2em}
    \caption{\textbf{Ablation of dictionary update.}
    WikiText-2 perplexity of LLaMA-7B across different sparsity levels, with and without dictionary update.
    }
    \vspace{-2em}
    \label{fig:mod_update}
\end{wrapfigure}

 \paragraph{{\tb Ablation of dictionary update.}}
We ablate the effect of updating the dictionary after coefficient pruning (Fig.~\ref{fig:mod_update}). 
We keep the sparse coefficient matrices fixed and refit the shared dictionary across a coefficient sparsity range from 20\% to 50\%.
This step improves the reconstruction obtained with the fixed sparse codes without changing the sparsity pattern or the model-size estimate.
The dictionary update reduces perplexity compared to the pruned baseline.
The interesting part is that perplexity reduction is higher at increased sparsity: 1.04 PPL at 50\% sparsity, whereas 0.06 PPL at 20\% sparsity.
This suggests that coefficient pruning yields a suboptimal dictionary for the remaining active atoms, and that the dictionary refit alleviates some of the resulting increase in perplexity.

\paragraph{Different quantization format.}
We investigate how different precision choices for the dictionary and coefficients affect the final compression trade-off (Table~\ref{tab:quant_ablation}).
The FP16/FP16 baseline corresponds to the factorized model with a 30\% reduction in factorizations and 50\% coefficient sparsity before quantization.
Quantizing only the dictionary to NVFP4 yields a clear storage gain with a small increase in perplexity, suggesting that the shared dictionary is relatively robust to low-bit quantization.
When both the dictionary and coefficients are quantized to NVFP4, perplexity remains close to the INT8-coefficient setting while using almost half the precision.
Although NVFP4/FP8 mixed quantization exhibits a decent perplexity-size trade-off, our all-NVFP4 setting yields the most compact representation and pushes the model-size frontier further to the left.

\begin{table}[t]
    \centering
    \small
    \caption{
    \textbf{Different quantization format.}
    We compare quantization choices for the shared dictionary $\mathbf{D}$ and layer-specific coefficients $\mathbf{C}$.
    Perplexity is measured on WikiText-2 using LLaMA-7B, and compression gain denotes the reduction in model size.
    The FP16/FP16 baseline corresponds to the factorized model with a 30\% factorization reduction and 50\% coefficient sparsity before quantization.
    }
    \label{tab:quant_ablation}
    \small
    \vspace{1em}
    \begin{tabular}{lcccc}
        \toprule
        Method & Dictionary $\mathbf{D}$ & Coefficient $\mathbf{C}$ & PPL $\downarrow$ & Comp. gain $\uparrow$ \\
        \midrule
        {Unquantized baseline}    & { \texttt{FP16}}  & {\texttt{FP16}}  & {12.98}           & {1.00$\times$} \\
        Mixed precision 1               & \texttt{NVFP4} & \texttt{FP16}  & \textbf{13.66}  & 1.61$\times$ \\
        Mixed precision 2               & \texttt{NVFP4} & \texttt{FP8}   & 14.71           & 2.60$\times$ \\
        Mixed precision 3               & \texttt{NVFP4} & \texttt{INT8}  & 16.05           & 2.60$\times$ \\
        All NVFP4 (Ours)                & \texttt{NVFP4} & \texttt{NVFP4} & 16.82           & \textbf{3.56$\times$} \\
        \bottomrule
    \end{tabular}
\end{table}

\begin{table}[t]
    \centering
    \small
    \begin{minipage}[t]{0.54\textwidth}
        \centering
        \footnotesize
        \caption{
        \textbf{Generalization across model families.}
        Perplexity ({$\downarrow$}) of three different LLMs, e.g., LLaMA 2-7B, OPT-6.7B, and Mistral-7B, under 1.25\texttimes{} compression on WikiText-2.
        }
        \label{tab:multi_model}
        \vspace{1em}
        \begin{tabular}{lccc}
        \toprule
        {Method} & LLaMA2-7B & OPT-6.7B  & Mistral-7B \\
        \midrule
        ASVD \cite{yuan2023asvd}            & 10.10 & 82.0 & 13.72 \\
        SVD-LLM \cite{wangsvd}              & 8.5   & 16.04 & 10.21 \\
        Basis Sharing \cite{wang2025basis}  & 7.70  & 11.79 &  7.57 \\
        \sysname (Ours)                     & \textbf{6.93} & \textbf{11.36}  & \textbf{6.46} \\
        \bottomrule
        \end{tabular}
    \end{minipage}
    \hfill
    \begin{minipage}[t]{0.44\textwidth}
        \centering
        \footnotesize
    
        \caption{
        \textbf{Effect of LoRA fine-tuning.}
        WikiText-2 perplexity ($\downarrow$) of LLaMA-7B with post-compression LoRA fine-tuning across compression gains.
        }
        \label{tab:lora}
        \vspace{1em}
        \begin{tabular}{ccc}
            \toprule
            {Comp. gain} & Basis Sharing \cite{wang2025basis} & \sysname \\
            \midrule
            1.25\texttimes{} & 7.09  & \textbf{6.42} \\
            1.43\texttimes{} & 8.21  & \textbf{7.54} \\
            1.67\texttimes{} & 10.44 & \textbf{7.91} \\
            2\texttimes{} & 10.95 & \textbf{8.71} \\
            \bottomrule
        \end{tabular}
    \end{minipage}
    \vspace{-1em}
\end{table}

\paragraph{Generalization across model families.} 
Table~\ref{tab:multi_model} reports WikiText-2 perplexity at 1.25\texttimes{} compression for OPT-6.7B, LLaMA~2-7B, and Mistral-7B.
DictLLM consistently outperforms ASVD, SVD-LLM, and Basis Sharing across all three architectures.
These results suggest that dictionary factorization and dictionary-aware Hessian pruning generalize across model families.

\begin{wrapfigure}{r}{0.4\textwidth}
    \vspace{-2em}
    \centering
    \includegraphics[width=1.0\linewidth]{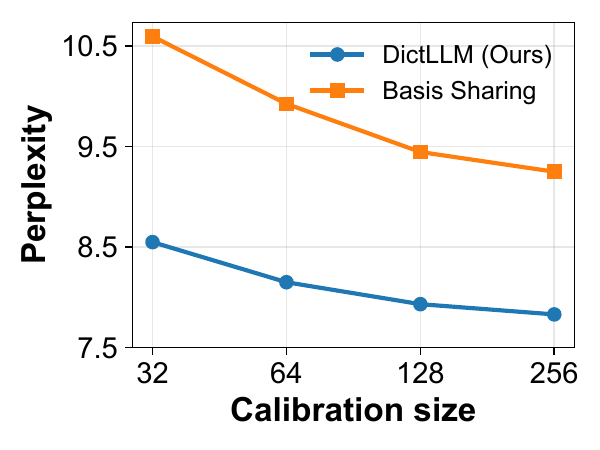}
    \vspace{-2em}
    \caption{\textbf{Effect of calibration size.}
    WikiText-2 perplexity of LLaMA-7B as a function of calibration set size.
    }
    \vspace{-2em}
    \label{fig:calib_length}
\end{wrapfigure}

\paragraph{Effect of LoRA fine-tuning.}
Table~\ref{tab:lora} reports the effect of applying low-rank adaptation (LoRA) fine-tuning~\cite{hu2022lora} after dictionary factorization and coefficient pruning.
LoRA fine-tuning consistently recovers a portion of the perplexity lost during compression, reducing WikiText-2 perplexity by up to 1.9 points.
The LoRA fine-tuning is more beneficial at higher compression ratios.
We report LoRA fine-tuning for both Basis Sharing and DictLLM updating only query and value weights.

\paragraph{Effect of calibration set size.}
We vary the number of calibration samples from 32 to 256 and measure WikiText-2 perplexity of LLaMA-7B at 1.43\texttimes{} compression (Fig.~\ref{fig:calib_length}).
DictLLM consistently outperforms Basis Sharing across all calibration sizes, while both DictLLM and Basis Sharing improve as more calibration data is available.
Notably, DictLLM, using only 32 samples, outperforms Basis Sharing with 256 samples.

\section{Conclusion} \label{sec:conclusion}

We present DictLLM, a post-training compression method that represents transformer weights with shared dictionary kernels and layer-specific sparse coefficients.
DictLLM first recovers shared dictionaries via activation-aware factorization, then prunes the coefficients using a dictionary-aware Hessian that projects the calibration covariance into the coefficient space.
We then refit the shared dictionary by using the resulting coefficients.
We further apply NVFP4 quantization with Hessian-based error compensation to both dictionary and coefficient factors, reducing storage while maintaining the factorized structure.
Comprehensive experiments show that DictLLM improves the perplexity-model size trade-off and preserves performance relative to state-of-the-art low-rank compression methods across language modeling and zero-shot reasoning benchmarks.

\begin{ack}
This work was supported in part by COGNISENSE, one of seven centers in JUMP 2.0, a Semiconductor Research Corporation (SRC) program sponsored by DARPA. 
The work of SY Chun was supported by IITP grants funded by the Korea government(MSIT) [No.RS-2021-II211343, Artificial Intelligence Graduate School Program (Seoul National University) / No.RS-2025-02314125, Effective Human-Machine Teaming With Multimodal Hazy Oracle Models].
\end{ack}


\medskip

\begingroup
{
\small
\bibliographystyle{unsrtnat}
\bibliography{reference}
}
\endgroup

\newpage
\clearpage
\appendix

\setcounter{section}{0}
\setcounter{figure}{0}
\setcounter{table}{0}
\setcounter{equation}{0}

\renewcommand{\thesection}{\Alph{section}}
\renewcommand{\thesubsection}{\thesection\arabic{subsection}}
\renewcommand{\thefigure}{A\arabic{figure}}
\renewcommand{\thetable}{A\arabic{table}}
\renewcommand{\theequation}{A\arabic{equation}}

\makeatletter
\begingroup
\renewcommand{\@seccntformat}[1]{}
\section[Appendix]{\centering\LARGE\bfseries Appendix}
\endgroup
\makeatother

\begingroup
\etocsetnexttocdepth{subsubsection}
\etocsettocstyle{\subsection*{Appendix contents}}{}
\localtableofcontents
\endgroup

\subsection{Additional details}
\label{app:add_detail}

\subsubsection{Additional details on dictionary learning}
\label{app:add_detail_dict}

\paragraph{Relation to \textit{k}-SVD.}
\label{app:ksvd_relation}
DictLLM is inspired by sparse dictionary learning, but its objective differs from classical \textit{k}-SVD \cite{aharon2006k}.
Given training signals $\mathbf{Y}=[\mathbf{y}_1,\ldots,\mathbf{y}_N]$, \textit{k}-SVD learns an overcomplete dictionary $\mathbf{D}\in\mathbb{R}^{d\times k}$ with $k>d$ and sparse codes $\mathbf{A}$ by solving
\begin{equation}
    \min_{\mathbf{D},\,\mathbf{A}}
    \left\|
    \mathbf{Y}-\mathbf{D}\mathbf{A}
    \right\|_{F}^{2}
    \quad
    \mathrm{s.t.}\quad
    \|\mathbf{a}_i\|_0 \leq s,\;\forall i.
\end{equation}
Although \textit{k}-SVD is often used with an overcomplete dictionary, our comparison uses an undercomplete dictionary ($k<d_\text{in}$) to match the compression setting.
\textit{k}-SVD is computationally expensive because it updates dictionary atoms and their associated sparse coefficients one at a time by repeatedly forming residuals and solving local rank-one approximation problems.

In contrast, DictLLM does not learn a dictionary from raw data signals.
Instead, it starts from pretrained weights and minimizes the activation-space reconstruction error induced by replacing each dense weight with a shared dictionary and sparse coefficients:
\begin{equation}
    \min_{\mathbf{D}^{g},\,\{\mathbf{C}^{\ell}\}_{\ell \in g}}
    \sum_{\ell\in g}
    \left\|
    \mathbf{X}^{\ell}\mathbf{W}^{\ell}
    -
    \mathbf{X}^{\ell}\mathbf{D}^{g}\mathbf{C}^{\ell}
    \right\|_{F}^{2}
    \quad
    \mathrm{s.t.}\quad
    \|\mathbf{C}^{\ell}_{:,j}\|_0 \leq (1-p)k.
\end{equation}
Here, $\mathbf{D}^{g}\in\mathbb{R}^{d_\text{in}\times k}$ is a compression dictionary whose size $k$ is chosen from the target parameter budget, rather than an overcomplete dictionary.
This makes the problem closer to post-training model compression than classical sparse coding.
Accordingly, DictLLM uses a scalable staged approximation: activation-aware SVD initializes the dictionary, Hessian-aware pruning chooses the sparse coefficient support, and a MOD-style update refits the dictionary to the fixed sparse coefficients.

\paragraph{Relation to Method of Optimal Directions.}
The Method of Optimal Directions (MOD) \cite{engan1999method} is a classical dictionary-learning algorithm that alternates between estimating sparse codes and updating the dictionary to better reconstruct training signals.
Given fixed sparse codes, MOD updates all dictionary atoms at once by solving a least-squares problem.
Our refitting step uses the same fixed-code dictionary update idea, but in a post-training model-compression setting.
Instead of learning a dictionary from raw signals, we start from pretrained transformer weights, keep the pruned coefficient matrices fixed, and refit the shared dictionary to reduce the reconstruction error of the original weights.
Thus, our update is a lightweight post-pruning refinement rather than a full alternating dictionary-learning procedure.




\subsubsection{Compression and number of parameters}
\label{app:compression_param}

\paragraph{Modeling number of parameters in LLMs with dictionary and coefficients.}
Let $\mathcal{T}$ denote the set of weight types, such as attention and MLP projection matrices.
Assuming that each weight $\mathbf{W}_{t}^{\ell} \in \mathbb{R}^{d_{\text{in},t} \times d_{\text{out},t}}$, the total number of parameters in the original model is
\begin{equation}
    N_{\mathrm{original}}
    =
    \sum_{\ell=1}^{L}
    \sum_{t\in\mathcal{T}}
    \left|
    \mathbf{W}_{t}^{\ell}
    \right|
    =
    \sum_{\ell=1}^{L}
    \sum_{t\in\mathcal{T}}
    d_{\mathrm{in},t}^{\ell}
    d_{\mathrm{out},t}^{\ell},
\end{equation}
where $\left|\mathbf{W}_{t}^{\ell}\right|$ denotes the number of scalar parameters in weight matrix $\mathbf{W}_{t}^{\ell}$. 
This count includes only the transformer projection matrices compressed by DictLLM.
It excludes the tokenizer, embedding tables, normalization parameters, biases, and the language modeling head, which are kept unchanged.
DictLLM’s grouped dictionary parameter count can naturally follow as
\begin{equation}
    N_{\mathrm{DictLLM}}
    =
    \sum_{t\in\mathcal{T}}
    \sum_{g\in\mathcal{G}_{t}}
    \left(
    d_{\mathrm{in},t}^{g} k_{t}^{g}
    +
    (1-p_{t}^{g}) k_{t}^{g}
    \sum_{\ell\in g}
    d_{\mathrm{out},t}^{\ell}
    \right).
\end{equation}


\paragraph{Compression ratio and sparsity.}
We first describe how factorization determines the dictionary rank for one group of weights.
Consider $n$ matrices with the same shape $d_{\mathrm{in}}\times d_{\mathrm{out}}$ that share one dictionary of rank $k$.
Before coefficient pruning, i.e., with sparsity $p=0$, the factorization compression ratio is
\begin{equation}
    \rho_{\mathrm{fac}}
    =  1-
    \frac{
    d_{\mathrm{in}}k
    +
    nkd_{\mathrm{out}}
    }{
    nd_{\mathrm{in}}d_{\mathrm{out}}
    }.
\end{equation}
Solving for $k$ gives
\begin{equation}
    k
    =
    \frac{
    n(1-\rho_{\mathrm{fac}})d_{\mathrm{in}}d_{\mathrm{out}}
    }{
    d_{\mathrm{in}} + nd_{\mathrm{out}}
    }.
\end{equation}
For example, the query projection in LLaMA-7B has $d_{\mathrm{in}}=d_{\mathrm{out}}=4096$.
With $\rho_{\mathrm{fac}}=20\%$ and a sharing group of $n=2$ layers, this gives $k=2184$.
In contrast, the down projection has shape $11008 \times 4096$ and is not shared across layers following Basis Sharing, so $n=1$ and $k=2388$.
We choose the dictionary rank separately for each weight type so that the factorization compression ratio is matched across weight types.

After the rank is fixed, we apply sparsity level $p$ to the coefficients, leaving only a $(1-p)$ fraction of coefficient entries nonzero.
The resulting final compression ratio is
\begin{equation}
    \rho =1-
    \frac{
    d_{\mathrm{in}}k + kd_{\mathrm{out}}n(1-p)
    }{
    d_{\mathrm{in}}d_{\mathrm{out}}n
    }.
\end{equation}
We apply the same coefficient sparsity across all weight types, so the resulting compression ratio can vary by weight type.
For instance, when pruning a LLaMA-7B model factorized with a 20\% reduction and layer sharing group size $n=2$, we use coefficient sparsity $p=0.194$ to reach 30\% overall compression.
Under this setting, the effective compression by weight type is 30.36\% for Q, K, and V projections, 33.10\% for up and gate projections, 27.78\% for output projections, and 24.21\% for down projections.

\subsubsection{Dictionary-induced coefficient Hessian.}
\label{app:coef_hessian}
Let $\widetilde{\mathbf{C}}^{\ell}=\mathbf{C}^{\ell}+\Delta\mathbf{C}^{\ell}$ be the pruned coefficient.
Since the dictionary is fixed, the induced weight perturbation is
$\Delta\mathbf{W}^{\ell}=\mathbf{D}^{g}\Delta\mathbf{C}^{\ell}$.
Under the calibration reconstruction objective, the local pruning loss is
\begin{equation}
    \left\|
    \mathbf{X}^{\ell}\Delta\mathbf{W}^{\ell}
    \right\|_{F}^{2}
    =
    \left\|
    \mathbf{X}^{\ell}\mathbf{D}^{g}\Delta\mathbf{C}^{\ell}
    \right\|_{F}^{2}
    =
    \operatorname{Tr}
    \left(
    (\Delta\mathbf{C}^{\ell})^{\top}
    (\mathbf{D}^{g})^{\top}
    (\mathbf{X}^{\ell})^{\top}\mathbf{X}^{\ell}
    \mathbf{D}^{g}
    \Delta\mathbf{C}^{\ell}
    \right).
\end{equation}

For a single layer, this derivation gives the coefficient-space Hessian
\begin{equation}
\mathbf{H}^{\mathbf{C},\ell}
=
(\mathbf{D}^{g})^{\top}
\mathbf{S}^{\ell}
\mathbf{D}^{g}
.
\end{equation}
In practice, we use a shared group-level Hessian proxy
$\mathbf{H}^{\mathbf{C},g}
=
(\mathbf{D}^{g})^{\top}
\mathbf{S}^{g}
\mathbf{D}^{g}$,
where $\mathbf{S}^{g}=\sum_{\ell\in g}\mathbf{S}^{\ell}$.
This pooled covariance approximation avoids computing and storing separate Hessian inverses for every layer while preserving activation-aware sensitivity at the group level.

\subsection{Additional experimental details}
\label{app:add_exp_detail}

Our comparison excludes language modeling heads and embeddings, following the protocol of prior work; we count the number of non-zero entries in the coefficient matrices.
Unless otherwise specified, we use damping $\lambda=10^{-4}$ when regularizing covariance and Hessian matrices before inversion or Cholesky decomposition.
We use a relative Hessian damping factor of 0.1 for masked quantization.
For LM Harness \cite{eval-harness} evaluation, we report \texttt{acc} as the accuracy of all reasoning tasks instead of \texttt{acc\_norm}.
For PTB \cite{marcus1993building} evaluation, we strip leading and trailing whitespace from each test sentence and join sentences with double newlines before tokenization, following our original evaluation protocol

\begin{table}[h]
    \centering
    \small
    \begin{minipage}[t]{0.44\textwidth}
        \centering
        \caption{\textbf{Factorization and coefficient sparsity settings for LLaMA-7B.}
         $\rho_{\text{fac}}$ denotes the parameter reduction from factorization, and sparsity denotes the fraction of pruned coefficient entries.
        }
        \label{tab:supp-sparsity-llama}
        \small
        \begin{tabular}{cccc}
            \toprule
            Compression & Factorization & Sparsity \\
            gain & ratio $\rho_{\text{fac}}$ & $p$ \\
            \midrule
            1.25\texttimes{} & 10\% & 17.25\% \\
            1.43\texttimes{} & 20\% & 19.40\% \\
            1.67\texttimes{} & 30\% & 22.17\% \\
            2\texttimes{}    & 30\% & 44.38\% \\
            \bottomrule
        \end{tabular}
    \end{minipage}
    \hfill
    \begin{minipage}[t]{0.54\textwidth}
        \centering
        \caption{\textbf{Coefficient sparsity settings across LLaMA2-7B, OPT-6.7B, and Mistral-7B.}
         All models have a factorization ratio $\rho_{\text{fac}} = 10\%$ and corresponding coefficient sparsities to achieve 1.25\texttimes{} compression gain.
        }
        \label{tab:supp-sparsity-rest}
        \small
        \vspace{1em}
        \begin{tabular}{cccc}
            \toprule
            {Models} & LLaMA2-7B & OPT-6.7B  & Mistral-7B \\
            \midrule
            Sparsity $p$ & 17.25\% & 19.42\% & 17.66\% \\
            \bottomrule
        \end{tabular}
    \end{minipage}
\end{table}

For the DictLLM results in Table~\ref{tab:benchmark}, we report the reduction from factorization and coefficient sparsity in Table~\ref{tab:supp-sparsity-llama}.
All compressed models are calibrated on WikiText-2 and evaluated without task-specific fine-tuning.
Quantized models DictLLM\textsubscript{FP4} reported at the bottom of Table~\ref{tab:benchmark} have $\rho_{\text{fac}}= \{ 20\%,30\%,30\% \}$ with sparsity $p=\{19.40\%, 22.17\%, 44.38\%\}$, achieving compression gains of 5.08\texttimes{}, 5.94\texttimes{}, and 7.12\texttimes{}.

In Table~\ref{tab:prune_target_ablation}, the dense-dense setting reaches 1.43\texttimes{} compression using factorization alone.
For the remaining settings, we start from a factorized model with $\rho_\text{fac} = 20 \%$ and apply pruning to the dictionary, the coefficients, or both, using the sparsity levels $p$ listed in the table.

In Table~\ref{tab:multi_model}, we use models factorized with $\rho_\text{fac} = 10\%$ and sparsify the layer-specific coefficients with sparsity of 17.25\% for LLaMA2-7B, 19.42\% for OPT-6.7B, and 17.66\% for Mistral-7B as described in Table~\ref{tab:supp-sparsity-rest}.
To reach 1.25\texttimes{} compression (or 20\% compression rate), the effective compression rate of LLaMA2-7B by weight type is 20.37\% for Q, K, and V projections, 23.09\% for up and gate projections, 17.77\% for output projections, and 14.23\% for down projections.
On the other hand, OPT-6.7B has per-weight type compression rates of 21.67\% for Q, K, and V projections, 25.56\% for the up projection, 18.75\% for the output projections, and 13.50\% for the down projections.
Mistral-7B compressed by 1.25\texttimes{} has compression rates of 20.61\% for Q, 13.21\% for K and V, 17.96\% for the output, 23.92\% for the gate and up, and 13.54\% for the down projections.

For the LoRA fine-tuning results in Table~\ref{tab:lora}, we fine-tune DictLLM models at final compression gains from 1.25\texttimes{} to 2\texttimes{}, using the same factorization and pruning hyperparameters reported in Table~\ref{tab:benchmark}.
We use rank $r=8$ adapters with a scaling factor $\alpha=32$, yielding an effective LoRA update scale of $\alpha/r=4$. 
We train only the LoRA adapter parameters while keeping the compressed backbone fixed. 
Fine-tuning is performed with a batch size of 1, a learning rate of $10^{-4}$, and 2 training epochs. 
These experiments use the factorized and pruned models without dictionary refitting.

In Figure~\ref{fig:mod_update}, we start from a model with a 20\% factorization reduction, apply coefficient pruning with sparsity levels from 20\% to 50\%, and then apply dictionary refitting to each setting.

For the calibration set length sweep in Figure~\ref{fig:calib_length}, we use the 1.43\texttimes{} compressed DictLLM model with factorization rate $\rho_\text{fac} = 20 \%$, coefficient sparsity $p=19.40\%$, and no dictionary refitting.

\subsection{Generalization across model families and scales}
\label{app:supp-qwen3}

\begin{table*}[h]
    \centering
    \captionof{table}{\textbf{
    Benchmark on language modeling and reasoning tasks with Qwen3-8B.}
    We compare DictLLM against low-rank compression baselines at compression gains from 1.25\texttimes{} to 2\texttimes{}.
    We report perplexity on WikiText-2 and C4, and zero-shot accuracy on OpenBookQA (OB), WinoGrande (WG), HellaSwag (HS), PIQA (PQ), MathQA (MQ), ARC-Easy (AE), and ARC-Challenge (AC). 
    }
    \label{tab:supp-qwen8}
    \small
    \setlength{\tabcolsep}{3pt}
    \resizebox{1\textwidth}{!}{%
    \begin{tabular}{c|c|cc|cccccccc}
        \toprule
        \textbf{Comp.} & \multirow{2}{*}{\textbf{Method}} & \multicolumn{2}{c|}{\textbf{Perplexity} $\downarrow$} & \multicolumn{8}{c}{\textbf{Zero-shot accuracy (\%)} $\uparrow$} \\
        \textbf{gain} & & WikiText-2 & C4 & OB & WG & HS & PQ & MQ & AE & AC & Average \\
        \midrule
        \midrule

        {\color{gray}1\texttimes{}} & {\color{gray}Original} & {\color{gray}7.09} & {\color{gray}11.87} & {\color{gray}32.20} & {\color{gray}72.53} & {\color{gray}59.03} & {\color{gray}79.16} & {\color{gray}55.41} & {\color{gray}81.61} & {\color{gray}53.24} & {\color{gray}61.88} \\
        \midrule

        \multirow{3}{*}{1.25\texttimes{}} & SVD-LLM~\cite{wangsvd} & 11.52 & 32.16 & 29.00 & 66.22 & 44.16 & 71.55 & 31.56 & 67.51 & 38.31 & 49.76 \\
        & Basis Sharing~\cite{wang2025basis} & 11.14 & 29.37 & \textbf{31.40} & 67.56 & 45.77 & 71.93 & 32.26 & 67.21 & 38.99 & 50.73 \\
        & DictLLM (Ours) & \textbf{9.90} & \textbf{22.59} & 31.00 & \textbf{70.09} & \textbf{50.07} & \textbf{74.86} & \textbf{35.78} & \textbf{70.54} & \textbf{41.89} & \textbf{53.46} \\
        \midrule

        \multirow{3}{*}{1.43\texttimes{}} & SVD-LLM~\cite{wangsvd} & 14.30 & 51.96 & 26.20 & 64.33 & 37.91 & 65.02 & 26.77 & 55.51 & 28.50 & 43.46 \\
        & Basis Sharing~\cite{wang2025basis} & 13.29 & 44.72 & 26.60 & 62.98 & 39.76 & 67.74 & 28.68 & 58.08 & 33.11 & 45.28 \\
        & DictLLM (Ours) & \textbf{11.26} & \textbf{30.09} & \textbf{31.60} & \textbf{67.64} & \textbf{45.21} & \textbf{72.36} & \textbf{32.56} & \textbf{65.66} & \textbf{38.82} & \textbf{50.55} \\
        \midrule

        \multirow{3}{*}{1.67\texttimes{}} & SVD-LLM~\cite{wangsvd} & 20.81 & 103.87 & 20.00 & 56.27 & 31.77 & 60.17 & 23.15 & 39.02 & 20.56 & 35.85 \\
        & Basis Sharing~\cite{wang2025basis} & 17.56 & 84.85 & 21.60 & 58.25 & 33.72 & 61.32 & 24.02 & 47.81 & 24.83 & 38.79 \\
        & DictLLM (Ours) & \textbf{13.59} & \textbf{47.79} & \textbf{26.40} & \textbf{63.22} & \textbf{39.05} & \textbf{67.03} & \textbf{27.71} & \textbf{58.08} & \textbf{32.94} & \textbf{44.92} \\
        \midrule

        \multirow{3}{*}{2\texttimes{}} & SVD-LLM~\cite{wangsvd} & 35.07 & 215.72 & 14.40 & 55.17 & 28.36 & 56.47 & 22.48 & 30.81 & 20.56 & 32.61 \\
        & Basis Sharing~\cite{wang2025basis} & 29.61 & 191.90 & 15.60 & 55.72 & 29.25 & 58.43 & 22.81 & 37.16 & 21.16 & 34.31 \\
        & DictLLM (Ours) & \textbf{16.58} & \textbf{73.52} & \textbf{24.20} & \textbf{60.14} & \textbf{35.38} & \textbf{63.06} & \textbf{26.37} & \textbf{54.04} & \textbf{28.24} & \textbf{41.63} \\
        \bottomrule
    \end{tabular}}
\end{table*}

\begin{table*}[h]
    \centering
    \captionof{table}{\textbf{Qwen3-1.7B language modeling and reasoning results.}
    Methods are compared at the reported compression gains.}
    \label{tab:supp-qwen17}
    \small
    \setlength{\tabcolsep}{3pt}
    \resizebox{1\textwidth}{!}{%
    \begin{tabular}{c|c|cc|cccccccc}
        \toprule
        \textbf{Comp.} & \multirow{2}{*}{\textbf{Method}} & \multicolumn{2}{c|}{\textbf{Perplexity} $\downarrow$} & \multicolumn{8}{c}{\textbf{Zero-shot accuracy (\%)} $\uparrow$} \\
        \textbf{gain} & & WikiText-2 & C4 & OB & WG & HS & PQ & MQ & AE & AC & Average \\
        \midrule
        \midrule

        {\color{gray}1\texttimes{}} & {\color{gray}Original} & {\color{gray}9.51} & {\color{gray}15.34} & {\color{gray}29.80} & {\color{gray}63.54} & {\color{gray}49.29} & {\color{gray}75.73} & {\color{gray}45.19} & {\color{gray}73.74} & {\color{gray}41.04} & {\color{gray}54.05} \\
        \midrule

        \multirow{3}{*}{1.25\texttimes{}} & SVD-LLM~\cite{wangsvd} & 23.20 & 96.37 & 22.80 & 56.83 & 34.98 & 61.64 & 24.15 & 44.15 & 24.66 & 38.46 \\
        & Basis Sharing~\cite{wang2025basis} & 22.93 & 95.07 & 22.60 & 56.20 & 36.66 & 62.02 & 26.16 & 48.82 & 27.13 & 39.94 \\
        & DictLLM (Ours) & \textbf{17.66} & \textbf{52.80} & \textbf{25.80} & \textbf{59.91} & \textbf{40.63} & \textbf{67.85} & \textbf{27.84} & \textbf{53.37} & \textbf{28.75} & \textbf{43.45} \\
        \midrule

        \multirow{3}{*}{1.43\texttimes{}} & SVD-LLM~\cite{wangsvd} & 34.60 & 190.49 & 19.00 & 55.01 & 31.08 & 58.32 & 23.75 & 38.13 & 22.95 & 35.46 \\
        & Basis Sharing~\cite{wang2025basis} & 33.37 & 206.33 & 21.00 & 55.09 & 32.44 & 58.60 & 23.79 & 42.63 & 23.38 & 36.70 \\
        & DictLLM (Ours) & \textbf{23.45} & \textbf{99.27} & \textbf{21.80} & \textbf{56.27} & \textbf{36.39} & \textbf{62.30} & \textbf{25.83} & \textbf{47.52} & \textbf{25.51} & \textbf{39.37} \\
        \bottomrule
    \end{tabular}}
\end{table*}

\begin{table*}[h]
    \centering
    \captionof{table}{\textbf{Qwen3-14B language modeling and reasoning results.}
    The improvements over other baselines extend to the largest model evaluated in language modeling and reasoning tasks.}
    \label{tab:supp-qwen14}
    \small
    \setlength{\tabcolsep}{3pt}
    \resizebox{1\textwidth}{!}{%
    \begin{tabular}{c|c|cc|cccccccc}
        \toprule
        \textbf{Comp.} & \multirow{2}{*}{\textbf{Method}} & \multicolumn{2}{c|}{\textbf{Perplexity} $\downarrow$} & \multicolumn{8}{c}{\textbf{Zero-shot accuracy (\%)} $\uparrow$} \\
        \textbf{gain} & & WikiText-2 & C4 & OB & WG & HS & PQ & MQ & AE & AC & Average \\
        \midrule
        \midrule

        {\color{gray}1\texttimes{}} & {\color{gray}Original} & {\color{gray}6.46} & {\color{gray}10.95} & {\color{gray}35.40} & {\color{gray}74.19} & {\color{gray}61.75} & {\color{gray}80.41} & {\color{gray}61.78} & {\color{gray}83.33} & {\color{gray}55.63} & {\color{gray}64.64} \\
        \midrule

        \multirow{3}{*}{1.25\texttimes{}} & SVD-LLM~\cite{wangsvd} & 9.49 & 27.15 & 28.80 & 68.98 & 47.19 & 72.69 & 34.34 & 67.00 & 38.48 & 51.07 \\
        & Basis Sharing~\cite{wang2025basis} & 9.18 & 25.93 & 30.20 & 70.24 & 49.32 & 74.32 & 36.98 & 66.08 & 39.33 & 52.35 \\
        & DictLLM (Ours) & \textbf{8.29} & \textbf{19.69} & \textbf{32.20} & \textbf{70.32} & \textbf{54.48} & \textbf{76.22} & \textbf{39.90} & \textbf{73.02} & \textbf{43.60} & \textbf{55.68} \\
        \midrule

        \multirow{3}{*}{1.43\texttimes{}} & SVD-LLM~\cite{wangsvd} & 11.51 & 44.54 & 26.00 & 65.11 & 39.29 & 67.08 & 29.95 & 62.25 & 31.57 & 45.89 \\
        & Basis Sharing~\cite{wang2025basis} & 10.81 & 40.77 & 28.80 & 68.67 & 42.67 & 70.24 & 32.83 & 60.35 & 35.07 & 48.38 \\
        & DictLLM (Ours) & \textbf{9.28} & \textbf{26.46} & \textbf{30.00} & \textbf{70.32} & \textbf{48.87} & \textbf{73.99} & \textbf{36.05} & \textbf{66.08} & \textbf{38.91} & \textbf{52.03} \\
        \bottomrule
    \end{tabular}}
\end{table*}  

Tables~\ref{tab:supp-qwen8}, \ref{tab:supp-qwen17}, and \ref{tab:supp-qwen14} compare DictLLM with SVD-LLM~\cite{wangsvd} and Basis Sharing~\cite{wang2025basis}. 
We evaluate 1.25\texttimes{} and 1.43\texttimes{} compression for all three scales, and extend the Qwen3-8B sweep to 1.67\texttimes{} and 2\texttimes{}. 
At every evaluated compression point, DictLLM achieves the lowest WikiText-2 and C4 perplexities and the highest average reasoning accuracy among these compression methods.
The results of SVD-LLM~\cite{wangsvd} and Basis Sharing~\cite{wang2025basis} are reproduced by us.

At 1.43\texttimes{} compression, DictLLM improves average accuracy over Basis Sharing from 36.70\% to 39.37\% on Qwen3-1.7B, from 45.28\% to 50.55\% on Qwen3-8B, and from 48.38\% to 52.03\% on Qwen3-14B. On Qwen3-8B at 2\texttimes{} compression, it reduces WikiText-2 perplexity from 29.61 to 16.58 and increases average accuracy from 34.31\% to 41.63\% relative to Basis Sharing.
These results extend the performance-model size trade-off to a newer model family and multiple parameter scales.

\begin{table}[h]
    \centering
    \caption{\textbf{Factorization and coefficient sparsity settings for Qwen3 model family.}
     We report the factorization ratio $\rho_{\text{fac}}$ and Sparsity $p$ required to achieve a given compression gain for Qwen3-8B, 1.7B, and 14B models.
    }
    \label{tab:supp-sparsity-qwen}
    \small
    \begin{tabular}{ccccc}
        \toprule
         {\multirow{2}{*}{\textbf{Compression gain}}} & {\multirow{2}{*}{\textbf{Factorization ratio} $\rho_{\text{fac}}$}} &  \multicolumn{3}{c}{\textbf{Sparsity} $p$} \\
         &  & Qwen3-8B & Qwen3-1.7B  & Qwen3-14B \\
        \midrule
        1.25\texttimes{} & 10\% & 17.25\% & 17.57\% & 17.52\%   \\
        1.43\texttimes{} & 20\% & 19.40\% & 19.79\% & 19.71\%   \\
        1.67\texttimes{} & 30\% & 22.17\% & -       & -         \\
        2\texttimes{}    & 30\% & 44.38\% & -       & -         \\
        \bottomrule
    \end{tabular}
\end{table}


We show the factorization ratio and coefficient sparsity to achieve a given compression gain for Qwen3 model families in Table~\ref{tab:supp-sparsity-qwen}

\subsection{Componentwise ablation}
\label{app:supp-pipeline}

\begin{table*}[h]
    \centering
    \caption{\textbf{Componentwise ablation of compression stages.}
    `F', `P', `R', and `Q' respectively denote factorization, Hessian pruning, dictionary refitting, and quantization.
    A checkmark indicates an active component.
    We measure language and reasoning performance on LLaMA-7B.
    }
    \label{tab:supp-pipeline}
    \footnotesize
    \setlength{\tabcolsep}{2pt}
    \resizebox{\textwidth}{!}{%
    \begin{tabular}{lc|*{4}{>{\centering\arraybackslash}p{1em}}|cc|cccccccc}
        \toprule
        \multirow{2}{*}{\textbf{Stage}} & \textbf{Comp.} & \multicolumn{4}{c|}{\textbf{Components}} & \multicolumn{2}{c|}{\textbf{Perplexity} $\downarrow$} & \multicolumn{8}{c}{\textbf{Zero-shot accuracy} $\uparrow$} \\
        & \textbf{gain} & F & P & R & Q & Wiki-2 & C4 & OB & WG & HS & PQ & MQ & AE & AC & Avg. \\
        \midrule
        {\tgr Original}     & 1\texttimes{} & -- & -- & -- & --  & {\tgr 5.68}    & {\tgr 7.34}      & {\tgr 0.28} & {\tgr 0.67} & {\tgr 0.56}  & {\tgr 0.78} & {\tgr 0.27}  & {\tgr 0.67}  & {\tgr 0.38} & {\tgr 0.52} \\
        + Factorization     & 1.43\texttimes{}  & $\checkmark$ & --             & --            & --            & 9.25  & 22.46 & 0.27 & 0.63 & 0.40 & 0.68 & 0.24 & 0.63 & 0.30 & 0.45 \\
        + Hessian pruning   & 2\texttimes{}     & $\checkmark$ & $\checkmark$   & --            & --            & 11.53 & 33.82 & 0.25 & 0.62 & 0.36 & 0.64 & 0.23 & 0.57 & 0.27 & 0.42 \\
        + Dictionary refit  & 2\texttimes{}     & $\checkmark$ & $\checkmark$   & $\checkmark$  & --            & 10.59 & 29.12 & 0.24 & 0.63 & 0.38 & 0.65 & 0.23 & 0.61 & 0.29 & 0.43 \\
        + Quantization & 7\texttimes{} & $\checkmark$ & $\checkmark$ & $\checkmark$ & $\checkmark$ & 11.92 & 33.90 & 0.24 & 0.59 & 0.35 & 0.63 & 0.22 & 0.54 & 0.26 & 0.40 \\

        \bottomrule
    \end{tabular}}
\end{table*}

Table~\ref{tab:supp-pipeline} evaluates the successive stages on LLaMA-7B under the same calibration and evaluation protocol.
Starting from factorization at $\rho_{\mathrm{fac}}=30\%$, coefficient-space Hessian pruning at approximately 44\% sparsity increases the reported compression gain from 1.43\texttimes{} to 2\texttimes{}, while average reasoning accuracy decreases from 0.45 to 0.42.
Dictionary refitting then raises accuracy to 0.43 and reduces WikiText-2 perplexity from 11.53 to 10.59 without changing the dictionary dimensions or coefficient sparsity.
C4 perplexity also improves, from 33.82 to 29.12.

Refitting, therefore, provides a quality recovery step without additional parameter storage, although it requires an additional offline update.
Subsequent quantization increases the reported compression gain to 7\texttimes{}, with WikiText-2 perplexity of 11.92 and an average accuracy of 0.40.
This cumulative comparison exposes the storage--quality trade-off at each stage; it is not a leave-one-out experiment at a fixed final budget.

\subsection{Spectral analysis and motivation for cross-layer weight sharing}
\label{app:spec_analysis}

\begin{figure}[h] 
	\centering
	   \includegraphics[width=1\linewidth]{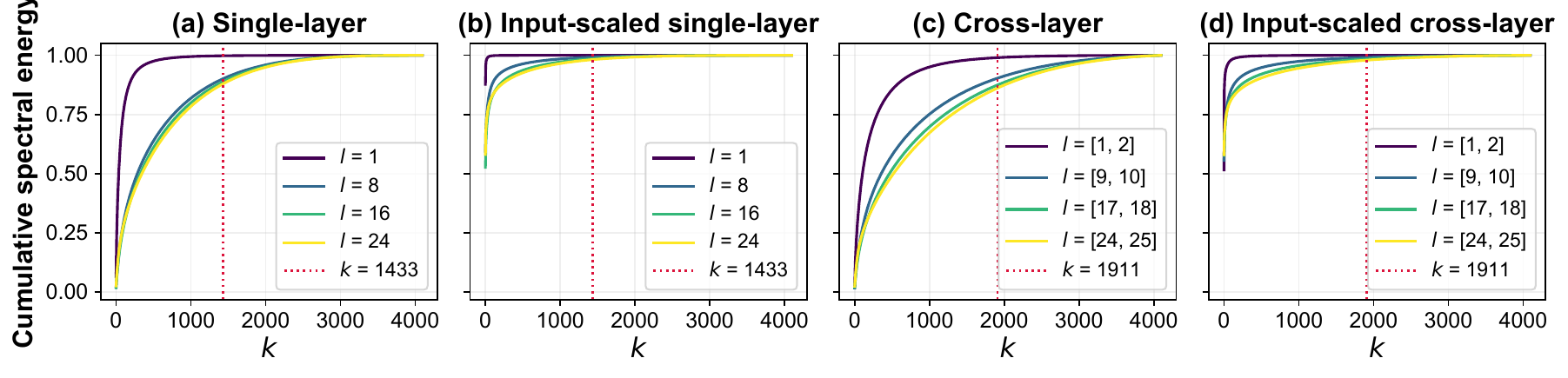}
        \caption{
        \textbf{Spectral concentration of LLaMA-7B query projection weights.} 
        We plot cumulative spectral energy for single-layer SVD (a) with and (b) without input covariance scaling and cross-layer SVD (c) with and (d) without scaling across layers $\ell$. 
        Input-scaled spectra rise more rapidly, showing that the energy of input-scaled weights is more concentrated in low-rank components than that of the raw weight spectrum.
        Vertical dashed lines indicate the selected rank $k \in \{1433, 1911\}$ for the target factorization rate $\rho_\text{fac} = 30 \%$ in single-layer and cross-layer settings.
        }
        \label{fig:svd_q}
\end{figure}

\paragraph{Motivation for input-scaled cross-layer factorization.}
We analyze the singular-value spectra of LLaMA-7B projection weights before and after cross-layer concatenation (\ref{fig:svd_q}).
For each setting, we compare the raw weight spectrum with the input-scaled spectrum computed using the WikiText-2 calibration covariance.
The input-scaled spectra accumulate faster, suggesting that fewer components are needed to preserve activation-weighted energy.
This supports calibration-aware factorization: while the raw weights may not be strongly low-rank, their action on calibration inputs is concentrated in a smaller set of spectral directions.

While cross-layer SVD has a smaller slope than the single-layer counterpart, it enables the shared dictionary to capture information common across adjacent layers, thereby easing the rank requirement $k = 1433 \rightarrow 1911$ for equal compression rate $\rho_\text{fac} = 30\%$.
We also observe that the spectrum is more concentrated in the early layers: for example, layer $\ell=1$ accumulates spectral energy rapidly at low ranks, whereas deeper layers, such as $\ell=24$, exhibit slower spectral decay.

\begin{figure}[h] 
	\centering
	\includegraphics[width=1\linewidth]{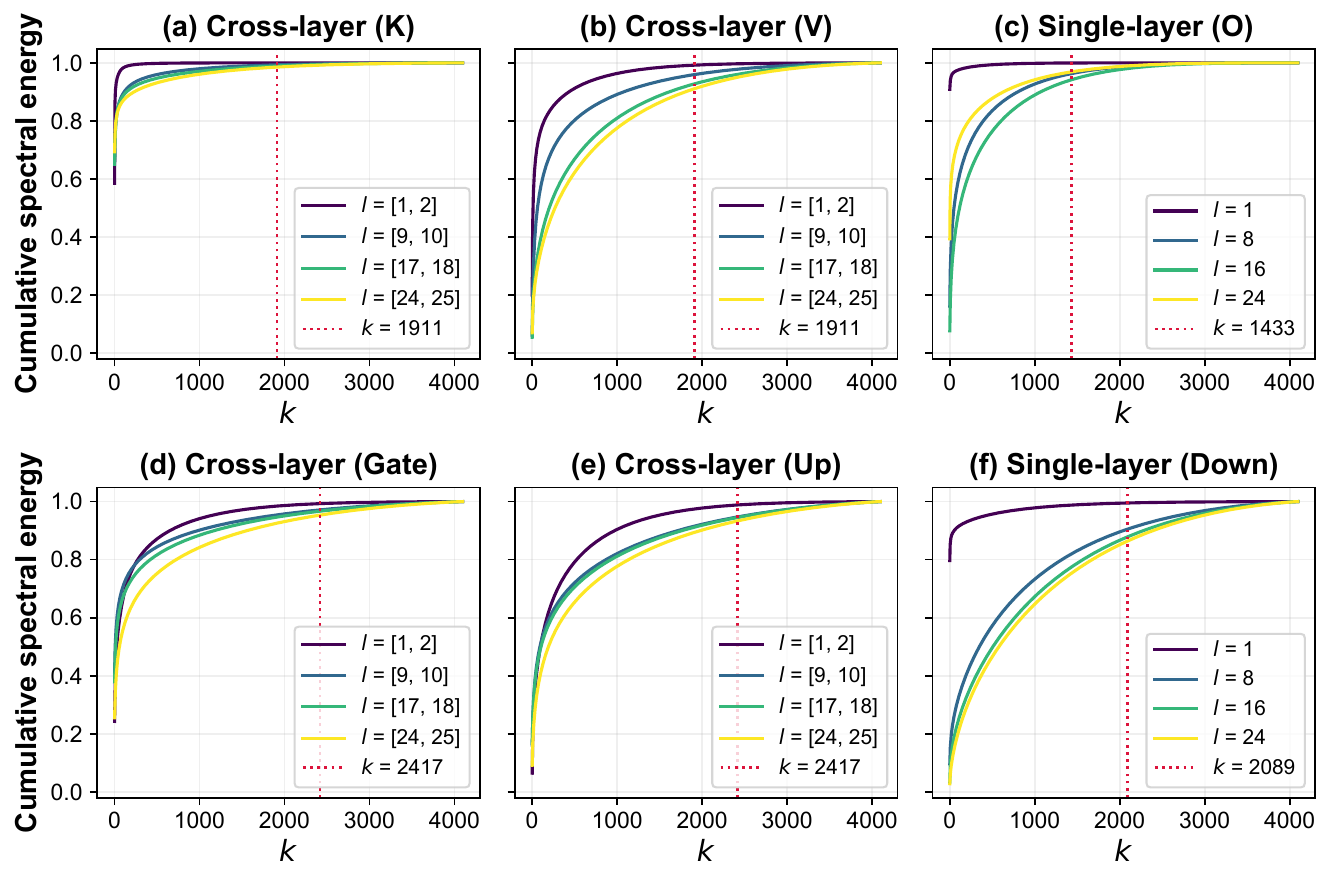}
        \caption{
        \textbf{Spectral concentration across different weight matrices in LLaMA-7B.}
        We plot cumulative spectral energy for the shared projections K, V, gate, and up using cross-layer SVD, and for the private projections O and down using single-layer SVD, following our factorization design.
        Vertical dashed lines indicate the selected rank $k$ for the target factorization rate $\rho_\text{fac}=30\%$ under the corresponding single-layer or cross-layer setting.
        }
        \label{fig:svd_others}
\end{figure}

\paragraph{Spectral analysis of shared and private projections.}
We separately analyze the projection families according to our sharing design and report the findings in Fig.~\ref{fig:svd_others}. 
For the shared projections K, V, gate, and up, we study the spectrum after cross-layer concatenation, since these matrices share a dictionary across layers.
In contrast, O and down are private projections in our model, so we analyze them with single-layer SVD. 

\begin{figure}[h] 
	\centering
	\includegraphics[width=0.7\linewidth]{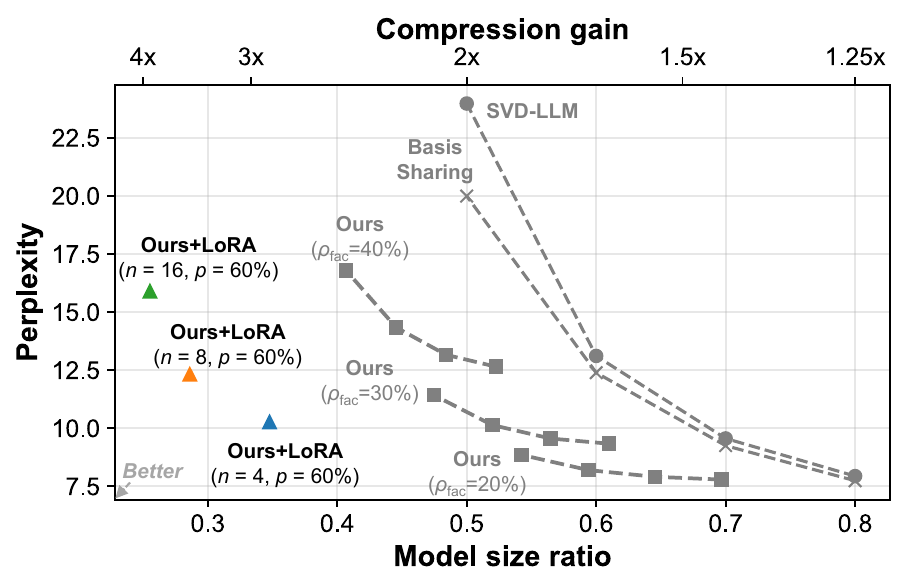}
        \caption{
        \textbf{Aggressive layer sharing with LoRA fine-tuning.}
        We increase the dictionary-sharing group size to $n = 4, 8, 16$ and apply 60\% coefficient sparsity while keeping the rank of the shared dictionary fixed, then use LoRA fine-tuning to restore performance.
        }
        \label{fig:layer_share_lora}
\end{figure}

\subsection{Aggressive layer sharing with LoRA fine-tuning.}
\label{app:layer-sharing}

\paragraph{Fine-tuning allows more agressive layer-sharing.}
Dictionary sharing reduces parameter count by amortizing a single dictionary across a group of layers, but increasing the sharing group beyond two layers can degrade performance~\cite[Table~6]{wang2025basis}.
To study a more aggressive compression setting, we also share the output and down projection matrices and increase the sharing group size to $n=4,8,16$.
We fix the shared dictionary ranks to $k_{\mathrm{q,k,v,o}}=2184$, $k_{\mathrm{up,gate}}=2762$, and $k_{\mathrm{down}}=3757$, corresponding to the ranks used at $\rho_{\mathrm{fac}}=20\%$.
We then prune the coefficients with sparsity $p=60\%$.
To recover the resulting perplexity degradation, we apply LoRA fine-tuning to all projection matrices using the same hyperparameters as in Table~\ref{tab:lora}.
The three settings achieve model-size reductions of 2.9\texttimes{}, 3.5\texttimes{}, and 3.9\texttimes{}, with WikiText-2 perplexities of 10.26, 12.32, and 15.89, respectively.

\begin{table}[h]
    \centering
    \footnotesize

    \caption{
    \textbf{Calibration size for NVFP4 quantization.}
    Post-quantization WikiText-2 perplexity of LLaMA-7B.
    Using 16 calibration samples provides a strong trade-off between perplexity and runtime.
    }
    \label{tab:quant_calib_length}
    \small
    \vspace{1em}
    \begin{tabular}{cc}
        \toprule
        Calibration size & Perplexity $\downarrow$ \\
        \midrule
        \textbf{16}  & 9.06 \\
        32           & 8.98 \\
        64           & 8.90 \\
        128          & \textbf{8.88} \\
        \bottomrule
    \end{tabular}
\end{table}

\paragraph{Ablation of calibration size on quantization.}
We apply NVFP4 quantization using subsets of the WikiText-2 training set for calibration.
Table~\ref{tab:quant_calib_length} shows that 16 calibration sequences provide an efficient accuracy-runtime trade-off.
Using 16 sequences achieves a perplexity of 9.06 in slightly over an hour, while increasing the calibration size to 128 takes 8\texttimes{} longer and improves perplexity by only 0.18 points.

\subsection{Comparison with additional baselines}

\begin{table*}[h]
    \centering
    \captionof{table}{\textbf{Comparison with pruning-only methods.}
    SparseGPT and Wanda are evaluated on LLaMA-7B at 80.32\% unstructured sparsity. 
    All results are reproduced by us.
    }
    \label{tab:supp-pruning}
    \small
    \setlength{\tabcolsep}{3pt}
    \resizebox{1\textwidth}{!}{%
    \begin{tabular}{c|c|cc|cccccccc}
        \toprule
        \textbf{Comp.} & \multirow{2}{*}{\textbf{Method}} & \multicolumn{2}{c|}{\textbf{Perplexity} $\downarrow$} & \multicolumn{8}{c}{\textbf{Zero-shot accuracy (\%)} $\uparrow$} \\
        \textbf{gain} & & WikiText-2 & C4 & OB & WG & HS & PQ & MQ & AE & AC & Average \\
        \midrule
        \midrule

        {\color{gray}1\texttimes{}} & {\color{gray}Original} & {\color{gray}5.68} & {\color{gray}7.34} & {\color{gray}30.80} & {\color{gray}69.38} & {\color{gray}56.36} & {\color{gray}78.24} & {\color{gray}26.57} & {\color{gray}74.66} & {\color{gray}41.55} & {\color{gray}53.94} \\
        \midrule

        {\multirow{3}{*}{5\texttimes{}}} & SparseGPT~\cite{frantar2023sparsegpt} & 86.72 & 257.96 & 14.00 & 48.46 & 25.38 & 51.41 & 18.79 & 25.63 & 21.08 & 29.25 \\
        {} & Wanda~\cite{sun2024simple} & 2.8~\texttimes{}~10\textsuperscript{3} & 2.8~\texttimes{}~10\textsuperscript{3} & 11.60 & 48.30 & 25.78 & 53.10 & 21.21 & 26.35 & 20.31 & 29.52 \\
        {} & {DictLLM\textsubscript{FP4}} & \textbf{8.48} & \textbf{17.04} & \textbf{29.20} & \textbf{64.64} & \textbf{43.67} & \textbf{70.18} & \textbf{23.99} & \textbf{64.02} & \textbf{32.76} & \textbf{46.92} \\

        \bottomrule
    \end{tabular}}
\end{table*}

\paragraph{Pruning-only baselines at high sparsity.}
Table~\ref{tab:supp-pruning} compares SparseGPT~\cite{frantar2023sparsegpt} and Wanda~\cite{sun2024simple} at 80.32\% unstructured sparsity with 4-bit DictLLM at a similar reported 5\texttimes{} gain.
DictLLM combines factorization at $\rho_{\mathrm{fac}}=20\%$ and approximately 19\% coefficient sparsity with quantization.
At this aggressive pruning level, SparseGPT and Wanda achieve average accuracies of 29.25\% and 29.52\%, respectively, whereas DictLLM achieves 46.92\%.
The language modeling results also favor DictLLM.
This comparison illustrates the benefit of distributing compression across factorization, sparsity, and precision, rather than relying solely on pruning to meet the budget.

\begin{table*}[h]
    \centering
    \captionof{table}{\textbf{Comparison with quantization-only baselines.}
    We compare methods at 7\texttimes{} compression gain on LLaMA-7B.
    GPTQ~\cite{frantar2023gptq} and AWQ~\cite{lin2024awq} use two-bit weights, whereas DictLLM combines factorization and coefficient sparsity with four-bit quantization.
    All results are reproduced by us.
    }
    \label{tab:supp-quantization}
    \small
    \setlength{\tabcolsep}{3pt}
    \resizebox{1\textwidth}{!}{%
    \begin{tabular}{c|c|cc|cccccccc}
        \toprule
        \textbf{Comp.} & \multirow{2}{*}{\textbf{Method}} & \multicolumn{2}{c|}{\textbf{Perplexity} $\downarrow$} & \multicolumn{8}{c}{\textbf{Zero-shot accuracy (\%)} $\uparrow$} \\
        \textbf{gain} & & WikiText-2 & C4 & OB & WG & HS & PQ & MQ & AE & AC & Average \\
        \midrule
        \midrule

        {\color{gray}1\texttimes{}} & {\color{gray}Original} & {\color{gray}5.68} & {\color{gray}7.34} & {\color{gray}30.80} & {\color{gray}69.38} & {\color{gray}56.36} & {\color{gray}78.24} & {\color{gray}26.57} & {\color{gray}74.66} & {\color{gray}41.55} & {\color{gray}53.94} \\
        \midrule

        \multirow{3}{*}{7\texttimes{}} & GPTQ~\cite{frantar2023gptq} (2-bit) & 17.59 & {36.56} & 13.80 & 48.38 & 30.47 & 57.13 & 20.84 & 32.03 & 21.33 & 32.00 \\
        {} & AWQ~\cite{lin2024awq} (2-bit) & 2.6~\texttimes{}~10\textsuperscript{5} & 2.9~\texttimes{}~10\textsuperscript{5} & 13.60 & 49.33 & 25.40 & 52.83 & 21.57 & 25.38 & 22.35 & 30.07 \\
        {} & {DictLLM\textsubscript{FP4}} & \textbf{11.92} & \textbf{33.90} & \textbf{23.60} & \textbf{59.19} & \textbf{35.32} & \textbf{63.17} & \textbf{22.21} & \textbf{53.62} & \textbf{25.60} & \textbf{40.39} \\
        
        \bottomrule
    \end{tabular}}
\end{table*}

\paragraph{Quantization-only baselines at a larger compression gain.}
Table~\ref{tab:supp-quantization} examines the more aggressive reported 7\texttimes{} budget.
NVFP4 supplies approximately a 3.56\texttimes{} reduction, including its scale overhead, which compounds with the approximately 2\texttimes{} gain from factorization and pruning at $\rho_{\mathrm{fac}}=30\%$ and approximately 44\% coefficient sparsity.
We therefore compare against 2-bit GPTQ~\cite{frantar2023gptq} and AWQ~\cite{lin2024awq}, rather than 4-bit quantization alone, to approach this larger compression gain.

DictLLM achieves an average accuracy of 40.39\%, compared with 32.00\% for GPTQ~\cite{frantar2023gptq} and 30.07\% for AWQ~\cite{lin2024awq}.
It also has lower WikiText-2 perplexity than either quantization-only baseline.

\begin{table}[h]
    \centering
    \small
    \centering
    \caption{\textbf{Comparison with low rank methods including SVD-LLM V2.}
    We measure WikiText-2 perplexity at 20\% compression.
    }
    \label{tab:supp-svdllm-v2}
    \setlength{\tabcolsep}{3pt}
    \begin{tabular}{lcc}
        \toprule
        Method & LLaMA-7B & OPT-6.7B \\
        \midrule
        ASVD \cite{yuan2023asvd}        & 11.14 & 82.0 \\
        SVD-LLM \cite{wangsvd}          & 7.94 & 16.04 \\
        SVD-LLM V2 \cite{wang2025svd}   & 7.12 & 13.46 \\
        Basis Sharing \cite{wang2025basis} & 7.74 & 11.79 \\
        DictLLM (Ours) & \textbf{6.89} & \textbf{11.36} \\
        \bottomrule
    \end{tabular}
\end{table}

\paragraph{Comparison against SVD-LLM V2.}
We supplement our comparisons with the results of SVD-LLM V2 \cite{wang2025svd}.
Table~\ref{tab:supp-svdllm-v2} reports WikiText-2 perplexities at 20\% compression on LLaMA-7B and OPT-6.7B.
DictLLM obtains perplexities of 6.89 and 11.36, compared with 7.12 and 13.46 for SVD-LLM V2.


\begin{table*}[h]
    \centering
    \captionof{table}{\textbf{Effect of compounding quantization on Basis Sharing.}
    `BS' denotes Basis Sharing~\cite{wang2025basis}.
    We measure language and reasoning performances on LLaMA-7B.
    All results are reproduced by us.
    }
    \label{tab:supp-bs-quant}
    \small
    \setlength{\tabcolsep}{3pt}
    \resizebox{1\textwidth}{!}{%
    \begin{tabular}{c|c|cc|cccccccc}
        \toprule
        \textbf{Comp.} & \multirow{2}{*}{\textbf{Method}} & \multicolumn{2}{c|}{\textbf{Perplexity} $\downarrow$} & \multicolumn{8}{c}{\textbf{Zero-shot accuracy (\%)} $\uparrow$} \\
        \textbf{gain} & & Wiki-2 & C4 & OB & WG & HS & PQ & MQ & AE & AC & Average \\
        \midrule
        \midrule

        {\color{gray}1\texttimes{}} & {\color{gray}Original} & {\color{gray}5.68} & {\color{gray}7.34} & {\color{gray}30.80} & {\color{gray}69.38} & {\color{gray}56.36} & {\color{gray}78.24} & {\color{gray}26.57} & {\color{gray}74.66} & {\color{gray}41.55} & {\color{gray}53.94} \\
        \midrule

        \multirow{3}{*}{5\texttimes{}} & BS \cite{wang2025basis} + GPTQ \cite{frantar2023gptq} (4-bit) & 13.99 & 34.01 & 24.00 & 56.83 {} & 34.72 & 57.89 & 20.74 & 46.93 & 23.46 & 37.80 \\
        {} &  BS \cite{wang2025basis} + AWQ \cite{lin2024awq} (4-bit) & 11.38 & 30.95 & 24.40 & 59.91 & 36.09 & 64.09 & 21.94 & 55.35 & 26.79 & 41.22 \\
        {} & {DictLLM\textsubscript{FP4}} & \textbf{8.48} & \textbf{17.04} & \textbf{29.20} & \textbf{64.64} & \textbf{43.67} & \textbf{70.18} & \textbf{23.99} & \textbf{64.02} & \textbf{32.76} & \textbf{46.92} \\

        \bottomrule
    \end{tabular}}
\end{table*}

\paragraph{Basis Sharing followed by quantization.}
We apply 4-bit GPTQ~\cite{frantar2023gptq} and AWQ~\cite{lin2024awq} to a Basis Sharing model factorized at $\rho_{\mathrm{fac}}=30\%$.
Table~\ref{tab:supp-bs-quant} compares these combinations with 4-bit DictLLM at the same reported 5\texttimes{} compression gain.
DictLLM uses $\rho_{\mathrm{fac}}=20\%$ and approximately 19\% coefficient sparsity before quantization.
It achieves WikiText-2 perplexity of 8.48, compared with 13.99 for Basis Sharing + GPTQ and 11.38 for Basis Sharing + AWQ.
Its average accuracy is 46.92\%, compared with 37.80\% and 41.22\%, respectively.
These results show that quantizing a factorized model alone does not match the quality of the DictLLM configuration at this reported budget.

\begin{table*}[h]
    \centering
    \captionof{table}{\textbf{Effect of compounding pruning and quantization on Basis Sharing.}
    `BS' denotes Basis Sharing~\cite{wang2025basis}.
    Magnitude pruning acts on the coefficient matrix.
    We measure language and reasoning performance on LLaMA-7B.
    All results are reproduced by us.
    }
    \label{tab:supp-bs-prune}
    \small
    \setlength{\tabcolsep}{3pt}
    \resizebox{1\textwidth}{!}{%
    \begin{tabular}{c|c|cc|cccccccc}
        \toprule
        \textbf{Comp.} & \multirow{2}{*}{\textbf{Method}} & \multicolumn{2}{c|}{\textbf{Perplexity} $\downarrow$} & \multicolumn{8}{c}{\textbf{Zero-shot accuracy (\%)} $\uparrow$} \\
        \textbf{gain} & & Wiki-2 & C4 & OB & WG & HS & PQ & MQ & AE & AC & Average \\
        \midrule
        \midrule

        {\color{gray}1\texttimes{}} & {\color{gray}Original} & {\color{gray}5.68} & {\color{gray}7.34} & {\color{gray}30.80} & {\color{gray}69.38} & {\color{gray}56.36} & {\color{gray}78.24} & {\color{gray}26.57} & {\color{gray}74.66} & {\color{gray}41.55} & {\color{gray}53.94} \\
        \midrule

        \multirow{2}{*}{1.43\texttimes{}} & BS~\cite{wang2025basis} + magnitude & 8.53 & 17.28 & \textbf{29.60} & 65.35 & 43.67 & 70.40 & 23.28 & 65.32 & \textbf{35.07} & 47.53 \\
        {} & {DictLLM} & \textbf{7.79} & \textbf{15.20} & {29.00} & \textbf{66.38} & \textbf{45.67} & \textbf{71.27} & \textbf{24.25} & \textbf{65.53} & \textbf{35.07} & \textbf{48.17} \\

        \midrule

        \multirow{2}{*}{5\texttimes{}} & BS~\cite{wang2025basis} + magnitude + AWQ~\cite{lin2024awq} (4-bit) & 10.39 & 23.50 & 23.80 & 62.98 & 38.64 & 66.05 & 22.28 & 57.32 & 25.51 & 42.37 \\
        {} & {DictLLM\textsubscript{FP4}} & \textbf{8.48} & \textbf{17.04} & \textbf{29.20} & \textbf{64.64} & \textbf{43.67} & \textbf{70.18} & \textbf{23.99} & \textbf{64.02} & \textbf{32.76} & \textbf{46.92} \\

        \bottomrule
    \end{tabular}}
    \vspace{-1em}
\end{table*}

\paragraph{Basis Sharing followed by magnitude pruning.}
To compare coefficient pruning strategies, we apply 19.4\% coefficient sparsity to a Basis Sharing model with $\rho_{\mathrm{fac}}=20\%$, and optionally apply 4-bit AWQ afterward.
As shown in Table~\ref{tab:supp-bs-prune}, DictLLM improves WikiText-2 perplexity from 8.53 to 7.79 and average accuracy from 47.53\% to 48.17\% before quantization.
With quantization, the corresponding values improve from 10.39 to 8.48 and from 42.37\% to 46.92\%.
The unquantized comparison supports the benefit of Hessian-based coefficient pruning together with dictionary refitting over magnitude-only coefficient pruning.
The quantized comparison also differs in its quantization method; therefore, it measures the combined compression pipeline rather than isolating pruning alone.

\subsection{End-to-end decoding throughput and memory}
\label{app:supp-deployment}

\begin{table}[h]
    \centering
    \caption{\textbf{End-to-end decoding on LLaMA-7B.}
    DictLLM uses MACKO for sparse coefficients.
    The measurement excludes the prompt prefill stage.
    }
    \label{tab:supp-macko-decode}
    \small
    \begin{tabular}{lccc}
        \toprule
        Configuration & Weight memory [GB] & Peak memory [GB] & Speed [tokens/s] \\
        \midrule
        Original & 13.48 & 13.66 & 102.35 \\
        Basis Sharing~\cite{wang2025basis} (dense $\mathbf{D}$, dense $\mathbf{C}$) & 9.59 & 9.83 & 113.59 \\
        {DictLLM} (dense $\mathbf{D}$, sparse $\mathbf{C}$) & \textbf{7.82} & \textbf{8.02} & \textbf{126.81} \\
        \bottomrule
    \end{tabular}
\end{table}

\paragraph{Benchmark setup.}
We use MACKO \cite{mackomacko} to perform sparse-coefficient operations and to measure whole-model decoding, rather than isolated matrix multiplication.
All configurations use FP16 precision on an NVIDIA RTX PRO 6000 Blackwell Max-Q GPU with a batch size of 1.
Each run uses a 128-token prompt and 100 single-token autoregressive decoding steps.
We report results over 10 measured runs following three warm-up runs.
Prompt prefill is excluded from the timed interval.

\paragraph{Memory saving and speed-up.}
Table~\ref{tab:supp-macko-decode} compares the dense LLaMA-7B model, Basis Sharing with $\rho_{\mathrm{fac}}=30\%$, and DictLLM with the same factorization rate and approximately 44\% coefficient sparsity.
DictLLM reaches 126.81 tokens/s, a 1.24\texttimes{} increase over the original model and a 1.12\texttimes{} increase over the factorization-only model.
Weight memory decreases from 13.48 to 7.82 GB (1.72\texttimes{} reduction), and peak memory decreases from 13.66 to 8.02 GB (1.70\texttimes{} reduction).
These results demonstrate a realized decoding benefit for the tested hardware and sparse backend, beyond the representation-level size estimates.


\subsection{Assets and compute resources}

\subsubsection{Assets}


\paragraph{License for WikiText-2.}
The WikiText-2~\cite{merity2017pointer} dataset is published under the CC-BY-SA-3.0 License, and all data\footnote{\url{https://huggingface.co/datasets/mindchain/wikitext2}} is released under the GNU Free Documentation License family.

\paragraph{License for PTB.}
The PTB~\cite{marcus1993building} dataset\footnote{\url{https://huggingface.co/datasets/ptb-text-only/ptb_text_only}} is released without an explicit license; however, the data is released under LDC User Agreement for Non-Members.

\paragraph{License for C4.}
The C4~\cite{raffel2020exploring} dataset\footnote{\url{https://huggingface.co/datasets/allenai/c4}} is released under the ODC-BY License.

\paragraph{License for LM Harness.}
The LM Harness framework~\cite{eval-harness} with code\footnote{\url{https://github.com/EleutherAI/lm-evaluation-harness}} is released under MIT License.

\paragraph{License for LLaMA-7B.}
The LLaMA-7B model~\cite{touvron2023llama} with a variant checkpoint\footnote{\url{https://huggingface.co/jeffwan/llama-7b-hf}} is licensed under License Non-commercial bespoke license.

\paragraph{License for LLaMA 2-7B.}
The LLaMA 2-7B model~\cite{touvron2023llama} with checkpoint\footnote{\url{https://huggingface.co/meta-llama/Llama-2-7b}} is licensed under Llama 2 Community License Agreement.

\paragraph{License for Mistral-7B.}
The Mistral-7B model~\cite{jiang2023mistral} with checkpoint\footnote{\url{https://huggingface.co/mistralai/Mistral-7B-v0.1}} is licensed under Apache License 2.0.

\paragraph{License for OPT-6.7B.}
The OPT-6.7B model~\cite{zhang2022opt} with checkpoint\footnote{\url{https://huggingface.co/facebook/opt-6.7b}} is licensed under OPT-175B License Agreement.

\paragraph{License for Qwen3.}
The Qwen 3 models~\cite{yang2025qwen3} with checkpoints for Qwen3-1.7B\footnote{\url{https://huggingface.co/Qwen/Qwen3-1.7B}}, -8B\footnote{\url{https://huggingface.co/Qwen/Qwen3-8B}}, and -14B\footnote{\url{https://huggingface.co/Qwen/Qwen3-14B}} are licensed under the Apache License  2.0.



\subsubsection{Computation resources.}
The experiments are conducted on a Red Hat Enterprise Linux 8 server equipped with dual NVIDIA RTX PRO 6000 Blackwell Max-Q GPUs, each equipped with 96GB VRAM, an AMD Ryzen Threadripper PRO 5975WX CPU @ 3.6-4.5GHz, and 256GB of RAM.

\subsection{Discussions}

\paragraph{Limitations.}
A limitation of DictLLM is that coefficient pruning produces unstructured sparsity.
While this reduces parameter storage and can be exploited with sparse kernels, its hardware efficiency depends on support for irregular sparse access patterns.
Exploring structured coefficient sparsity, such as block sparsity, is a promising direction for improving hardware efficiency and co-design.

\paragraph{Broader impact.}
DictLLM aims to reduce the memory footprint of large language models, making capable models more accessible on resource-constrained hardware and lowering the energy cost of inference.
This may broaden access to LLM deployment beyond large datacenters and enable more private, local, and low-latency applications.
At the same time, easier deployment of compressed LLMs can also lower the barrier to misuse, including spam generation, misinformation, or unauthorized automation.
Since DictLLM preserves the behavior of the underlying pretrained model as closely as possible, it may also retain that model's biases, safety risks, and failure modes.






\end{document}